\documentclass[runningheads]{llncs}

\usepackage[T1]{fontenc}

\usepackage{amsmath}

\usepackage{textcomp}

\usepackage{booktabs}
\usepackage{multirow}

\usepackage[bookmarks=false]{hyperref}

\usepackage{subcaption}

\usepackage[usenames, dvipsnames]{xcolor}
\usepackage{pgfplots}

\usetikzlibrary{calc, positioning, shapes.geometric}

\pgfplotsset{compat=newest}

\newcommand{\rpm}{\raisebox{.2ex}{\(\scriptstyle\pm\)}}

\begin{document}
\title{Learning to Estimate Two Dense Depths from LiDAR and Event Data\thanks{Supported in part by the Hauts-de-France Region and in part by the SIVALab Joint Laboratory (Renault Group---Universit\'e de technologie de Compi\`egne (UTC)---Centre National de la Recherche Scientifique (CNRS)).}}
%
%

\author{Vincent Brebion\orcidID{0000-0001-7429-0547} \and
Julien Moreau\orcidID{0000-0001-5008-9232} \and
Franck Davoine\orcidID{0000-0002-8587-6997}}
\authorrunning{V. Brebion et al.}
\institute{Heudiasyc (Heuristics and Diagnosis of Complex Systems) Laboratory, CNRS, Universit\'e de technologie de Compi\`egne (UTC), 60319 Compi\`egne Cedex, France
\email{\{vincent.brebion,julien.moreau,franck.davoine\}@hds.utc.fr}}
%
\maketitle              
\begin{abstract}
Event cameras do not produce images, but rather a continuous flow of events, which encode changes of illumination for each pixel independently and asynchronously. While they output temporally rich information, they lack any depth information which could facilitate their use with other sensors. LiDARs can provide this depth information, but are by nature very sparse, which makes the depth-to-event association more complex. Furthermore, as events represent changes of illumination, they might also represent changes of depth; associating them with a single depth is therefore inadequate. In this work, we propose to address these issues by fusing information from an event camera and a LiDAR using a learning-based approach to estimate accurate dense depth maps. To solve the ``potential change of depth'' problem, we propose here to estimate two depth maps at each step: one ``before'' the events happen, and one ``after'' the events happen. We further propose to use this pair of depths to compute a depth difference for each event, to give them more context. We train and evaluate our network, ALED, on both synthetic and real driving sequences, and show that it is able to predict dense depths with an error reduction of up to 61\% compared to the current state of the art. We also demonstrate the quality of our 2-depths-to-event association, and the usefulness of the depth difference information. Finally, we release SLED, a novel synthetic dataset comprising events, LiDAR point clouds, RGB images, and dense depth maps.

\end{abstract}
\section{Introduction}\label{sec:intro}
Rather than accumulating light to create images, event cameras perceive changes of illumination for each pixel independently and asynchronously. Thanks to their high temporal resolution (in the order of the \textmu{}s) and high dynamic range, event cameras are a sensor of choice for dynamic applications in complex environments (fast motions, extreme illumination conditions), where traditional cameras reach their limits.

LiDAR sensors offer accurate but sparse 3D information of their surrounding environment. They are a key component for autonomous navigation, helping to solve multiple problems, e.g., obstacle detection and tracking, SLAM, etc. Yet, their sparsity often constitutes a limiting factor. While 64- or 128-channel LiDARs are starting to be commercialized, they come at a significantly high cost, and are still not as dense as cameras.

In this work, we focus on the fusion of LiDAR and event data, which we consider as a dual problem: (1)~LiDAR depths densification and (2)~events-depths association. Regarding problem~(1), we are interested in densifying the LiDAR data using the events as a guide. As a result, dense depth maps are obtained, which allow for a dense 3D perception of the observed scene. As for problem~(2), we are interested in associating a depth to each event. By doing so, each event can be projected in 3D, and then even be backprojected in 2D in another vision sensor. For a fully calibrated and synced setup, this process would allow for the superimposition of events and RGB images, a task which is only possible at the moment through the use of specific low-resolution frame+events cameras like the DAVIS240C~\cite{Brandli2014A2}.

Estimating dense depth maps from sparse LiDAR data is a well-regarded problem as it solves the sparsity drawback of the LiDAR while keeping its metric scale. However, using events to densify depth maps (i.e., problem~(1)) might be inaccurately seen as a task that inherently includes problem (2), as corresponding depths for the events could be taken from the dense depth map. We argue in this work that, as each event represents a change in illumination, it might also represent a change in depth. As such, two depths should be associated to each event, and we will therefore compute two depth maps: one before the events happen, and one after they happen.

As an answer to these issues, we propose in this work a learning-based fusion method for estimating pairs of dense depth maps from events and sparse LiDAR data. Our main contributions are as follows:
\begin{itemize}
  \item We propose to revise the principle of associating a single depth to each event to rather estimate two depths: one ``before'' the event happens, and one ``after''. Following this, we introduce the notion of ``depth change map'', to give more context to each event.
  \item We propose a novel convolutional network, the ALED (Asynchronous LiDAR and Events Depths densification) network, able to fuse asynchronous events and LiDAR data, and to estimate the two dense depth maps from them, while surpassing state-of-the-art accuracy.
  \item We finally build and share a high-definition simulated dataset, the SLED (Synthetic LiDAR Events Depths) dataset, used as part of the training of the network and its evaluation.
\end{itemize}

If the reader is interested, supplementary material, the SLED dataset, source codes, as well as videos showcasing results on both simulated and real data are all available at \url{https://vbrebion.github.io/ALED}.

\section{Related Work}

\subsection{LiDAR Densification}
LiDAR sensors only produce sparse point clouds, which is challenging for numerous applications (3D reconstruction, object detection, SLAM, etc). As a consequence, LiDAR depth completion is a subject that has been widely studied in the literature.

Some authors try to obtain dense depth maps while only relying on the sparse data from the LiDAR. These methods either use machine learning~\cite{Chodosh2018DeepCC,Huang2020HMSNetHM,Uhrig2017SparsityIC} or traditional image processing operations~\cite{Ku2018InDO}.

The most successful approaches use a secondary modality as a guide for the densification process. While most of these approaches employ a RGB camera as the secondary sensor~\cite{VanGansbeke2019SparseAN,Huang2020HMSNetHM,Jaritz2018SparseAD,Xu2019DepthCF}, other authors have proposed using alternative modalities, such as stereo cameras~\cite{Maddern2016RealtimePF} or more recently event cameras~\cite{Cui2022DenseDE}.

\subsection{Fusion of Events and Other Modalities}
Due to their relative youth, the literature on the fusion of data from event cameras with other sensors is quite sparse.

Most of the investigations focused on the fusion of events and frames, thanks to sensors offering both modalities like the DAVIS camera~\cite{Brandli2014A2}. These works include frame interpolation and deblurring~\cite{Paikin2021EFINetVF,Pan2019BringingAB,Scheerlinck2018ContinuoustimeIE}, feature tracking~\cite{Gehrig2019EKLTAP,Kueng2016LowlatencyVO}, object detection~\cite{Cao2021FusionBasedFA,Jiang2019MixedFF,Tomy2022FusingEA}, or even steering prediction~\cite{Hu2020DDD20EE}.

In the past few years, a few authors have started investigating the fusion of events and LiDAR data. Explored issues include calibration~\cite{Song2018CalibrationOE,Ta2022L2ELT}, and very recently, point clouds enhancement with events~\cite{Li2021Enhancing3L} and LiDAR densification~\cite{Cui2022DenseDE}.

\subsection{Depth Estimations with Events}
Several approaches have been proposed in order to estimate sparse or dense depth maps by using a single event camera. Kim et al.~\cite{Kim2016RealTime3R} used probabilistic filters to simultaneously estimate the motion of the camera, reconstruct a log intensity image of the observed scene, and construct a sparse inverse depth map. Zhu et al.~\cite{Zhu2019UnsupervisedEL} used a convolutional network to jointly predict depth and egomotion, by trying to minimize the amount of motion blur in the accumulated events. Hidalgo-Carri\'o et al.~\cite{HidalgoCarrio2020LearningMD} were the first to estimate dense depth maps from a monocular event camera, through the use of a recurrent convolutional network.

In parallel, other authors have advocated for the use of a secondary sensor to help the depth estimation. Schraml et al.~\cite{Schraml2016AnES,Schraml2010DynamicSV} and Nam et al.~\cite{Nam2022StereoDF} used two event cameras, and estimated depths by creating images of accumulated events for each camera and applying stereo matching. While~\cite{Schraml2016AnES,Schraml2010DynamicSV} used traditional model-based approaches, \cite{Nam2022StereoDF} entirely relied on learning-based networks: an attention-based network to construct detailed events representations, then a convolutional network for depth map inference. Other authors have also combined the event camera with a RGB sensor; Gehrig et al.~\cite{Gehrig2021CombiningEA} for instance designed a recurrent network to fuse asynchronous data and estimate dense depths from them. Finally, some authors have also used depth sensors in direct combination with event cameras. Weikersdorfer et al.~\cite{Weikersdorfer2014Eventbased3S} used an RGB-D camera to obtain dense depths, and used the depth-augmented events to perform SLAM. Li et al.~\cite{Li2021Enhancing3L} used a LiDAR sensor to associate a depth to each event through the use of a Voronoi diagram and a set of heuritic rules. Cui et al.~\cite{Cui2022DenseDE} also employed a LiDAR, to derive dense depth maps by using 3D geometric information.

\section{Depth Change Map: Two Depths per Event}\label{sec:two_depths_per_event}

The fusion of depths and events is a problem that can have two different goals: (1)~obtaining dense depth maps from events and sparse LiDAR data, or (2)~determining a depth for each event. While problem~(1) can be interpreted as a LiDAR densification method guided by the events, we argue here for problem~(2) that associating a single depth to an event is inadequate.

By definition, an event represents a significant change in illumination observed by a given pixel. Under motion, observed events can either originate from (a)~texture changes inside an object; or from (b)~the contour of an object. In case~(a), associating a single depth to these events can be coherent, as depth inside an object should be subject to little variation. However, doing so in case~(b) is erroneous, as the events are likely to denote also a depth change.

Instead, we propose to estimate two dense depth maps: one before the events happen, which we will denote \(D_\text{bf}\) in the rest of this article, and one after the events happen, which we will denote \(D_\text{af}\). We can then formulate the depth change map as \(D_\text{af}-D_\text{bf}\) and compare the two depths \(d_\text{bf}\) and \(d_\text{af}\) for each pixel. Three meaningful cases can be distinguished:
\begin{enumerate}
  \item \(d_\text{af} - d_\text{bf} \approx 0\): the pixel is located in an area where depths do not vary much, i.e., inside an object;
  \item \(d_\text{af} - d_\text{bf} \gg 0\): the pixel was at the edge of an object, and is now on an object further away;
  \item \(d_\text{af} - d_\text{bf} \ll 0\): the pixel was located on a far object, and is now at the edge of a closer object.
\end{enumerate}

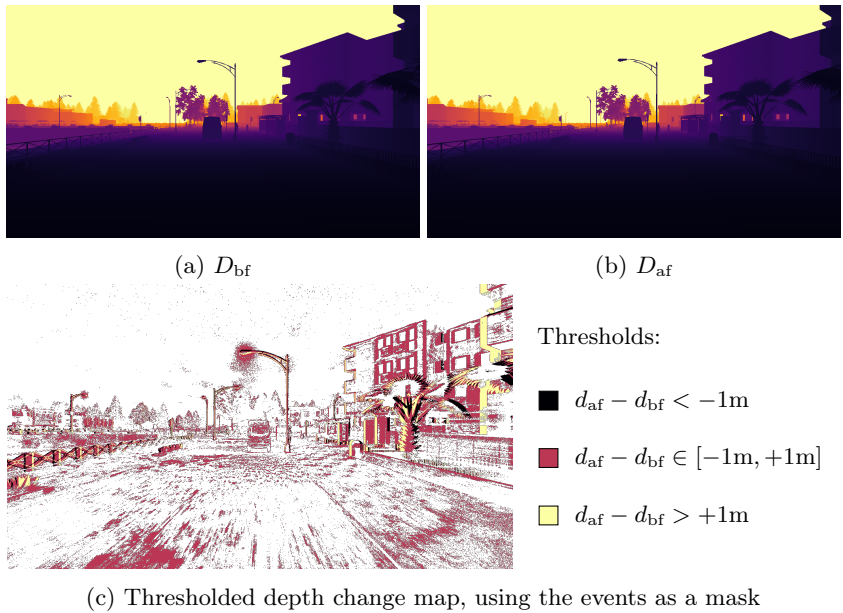
\begin{figure}[ht]
  \centering
  \begin{subfigure}{0.45\linewidth}
    \centering
    \includegraphics[width=\textwidth]{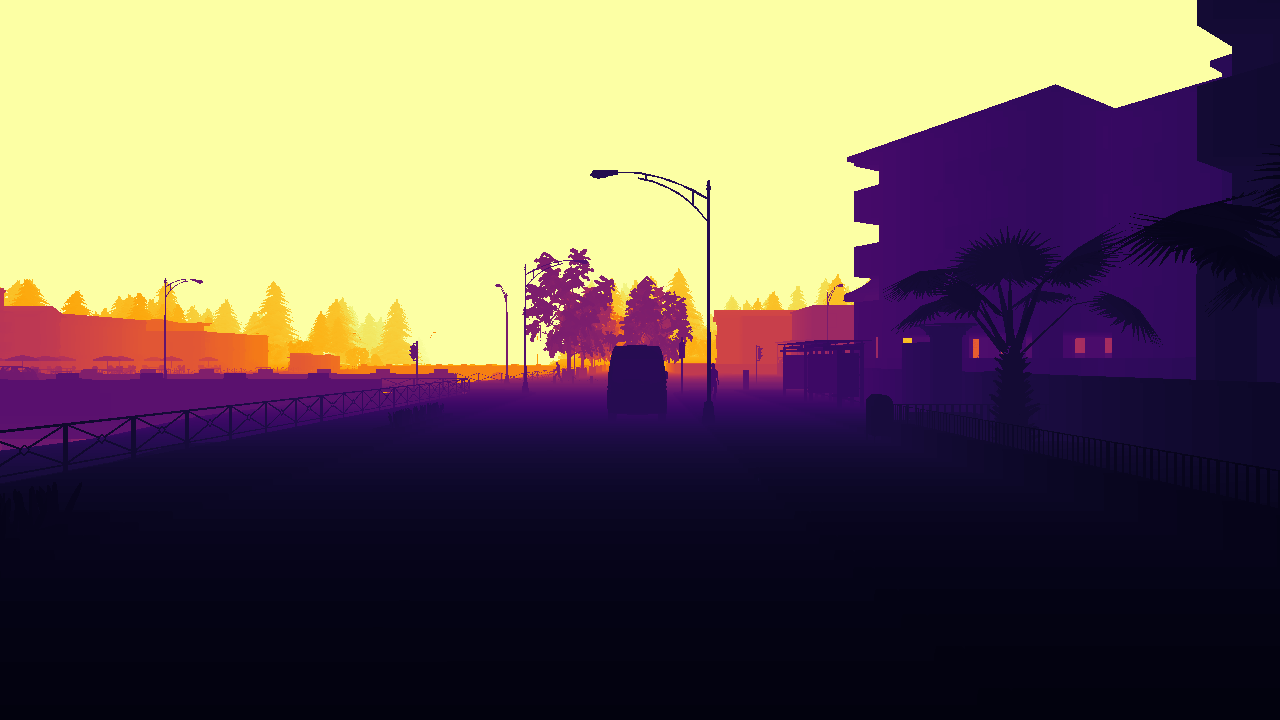}
    \caption{\(D_\text{bf}\)}
  \end{subfigure}
  \begin{subfigure}{0.45\linewidth}
    \centering
    \includegraphics[width=\textwidth]{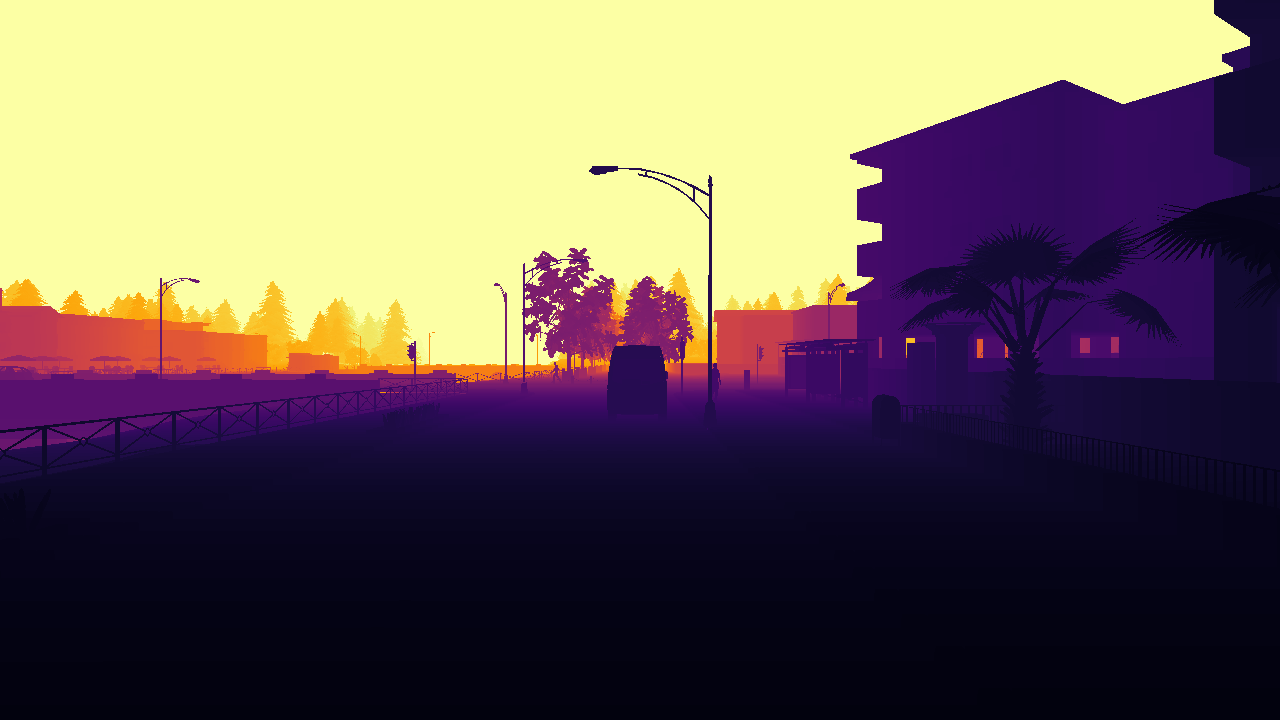}
    \caption{\(D_\text{af}\)}
  \end{subfigure}
  \begin{subfigure}{\linewidth}
    \centering
    \raisebox{-0.5\height}{\includegraphics[width=0.55\textwidth]{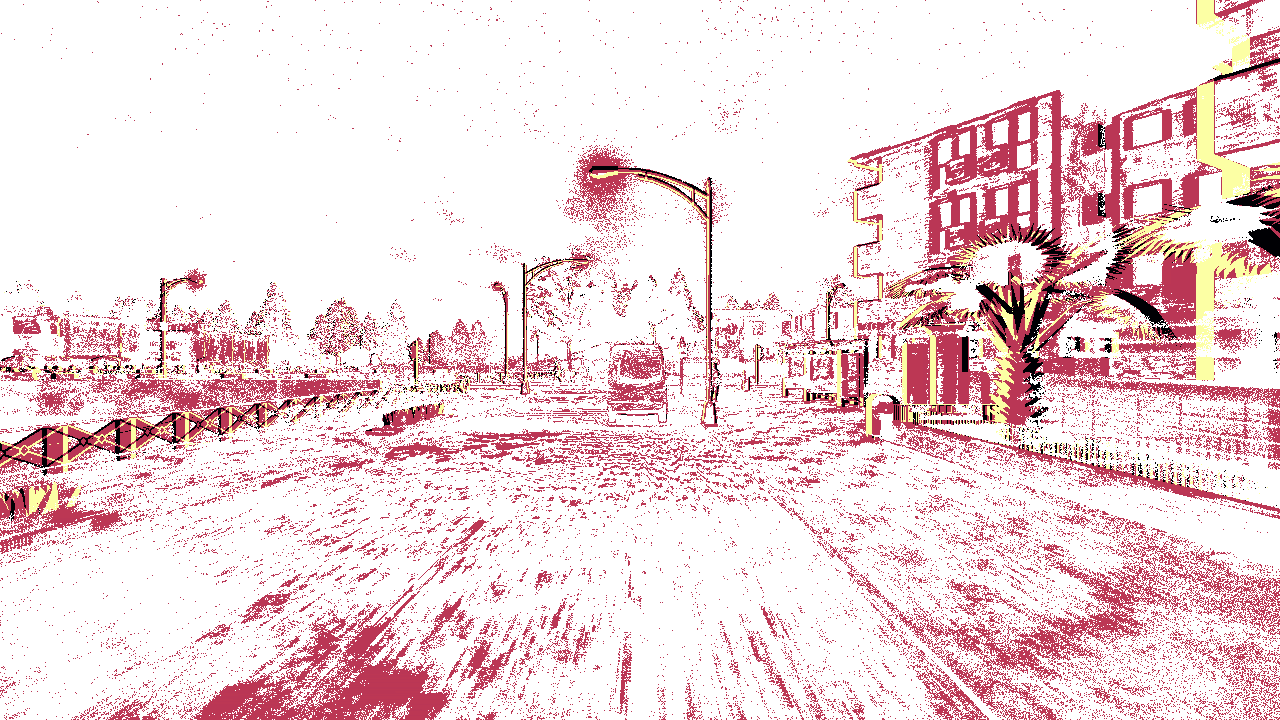}}
    \raisebox{-0.5\height}{
      \begin{tikzpicture}
        \definecolor{colorblack}{HTML}{000004}
        \definecolor{coloryellow}{HTML}{FCFFA4}
        \definecolor{colorpink}{HTML}{BC3754}
        \node[] (title) {Thresholds:};
        \node[draw, minimum width=0.25cm, minimum height=0.25cm, fill=colorblack, below left=0.5cm and -0.4cm of title] (black) {};
        \node[draw, minimum width=0.25cm, minimum height=0.25cm, fill=colorpink, below=0.5cm of black] (pink) {};
        \node[draw, minimum width=0.25cm, minimum height=0.25cm, fill=coloryellow, below=0.5cm of pink] (yellow) {};
        \node[right=0.1cm of black] (black_l) {\(d_\text{af} - d_\text{bf} < -1\text{m}\)};
        \node[right=0.1cm of pink] (pink_l) {\(d_\text{af} - d_\text{bf} \in [-1\text{m}, +1\text{m}]\)};
        \node[right=0.1cm of yellow] (yellow_l) {\(d_\text{af} - d_\text{bf} > +1\text{m}\)};
      \end{tikzpicture}
    }
    \caption{Thresholded depth change map, using the events as a mask}
  \end{subfigure}
  \caption{Example of the importance of the depth change map for each event on the ``Town01\_00'' sequence from our SLED dataset. Notice how simple thresholds on this depth difference help distinguishing the events linked to the contour of real objects from the events corresponding to the texture of the road, the halo from the street lamp, or even the noisy events in the sky.}\label{fig:depth_difference_example}
\end{figure}

The depth difference information given by the depth change map can especially help events processing to differentiate real objects from artifacts such as shadows and even noise. An illustration of some possibilities offered by the depth change map on events is given in Fig.~\ref{fig:depth_difference_example}. Other applications could also take advantage of the pair of depth maps \(D_\text{bf}\) and \(D_\text{af}\): ego-motion and speed estimation, objects clustering, scene flow, etc.

\section{Method}

\subsection{The ALED Network}

\begin{figure}
  \centering
  \resizebox{0.65\textwidth}{!}{
    \begin{tikzpicture}
      \node[] (L) at (0,0) {\includegraphics[width=0.2\textwidth]{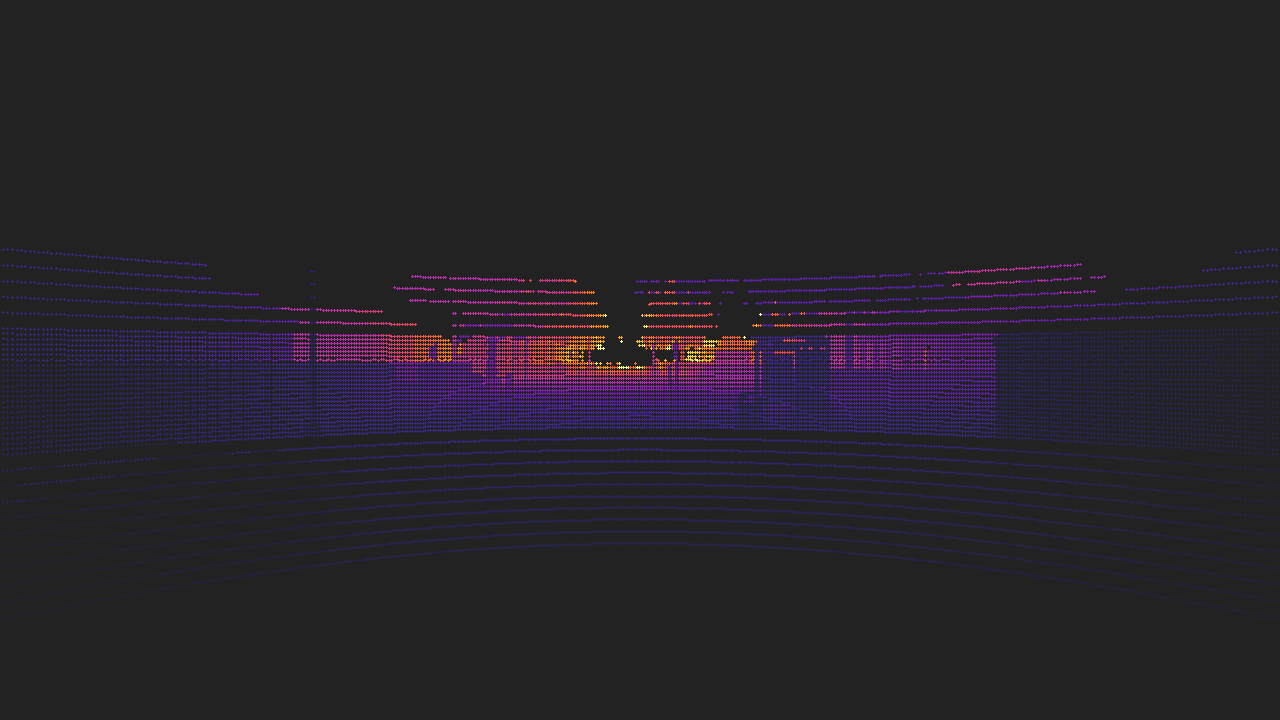}};
      \node[above=0cm of L] (Ll) {LiDAR};

      \node[draw, trapezium, minimum width=1cm, right= of L, opacity=0.0] (tmplh) {};
      \node[draw, trapezium, minimum width=1cm, fill=ForestGreen!40, rotate around={-90:(tmplh.center)}] (LH) at (tmplh) {};

      \node[draw, trapezium, minimum width=1cm, right= of tmplh, opacity=0.0] (tmple1) {};
      \node[draw, trapezium, minimum width=1cm, fill=yellow!40, rotate around={-90:(tmple1.center)}] (LE1) at (tmple1) {};

      \node[draw, trapezium, minimum width=1cm, right= of tmple1, opacity=0.0] (tmple2) {};
      \node[draw, trapezium, minimum width=1cm, fill=yellow!40, rotate around={-90:(tmple2.center)}] (LE2) at (tmple2) {};

      \node[draw, trapezium, minimum width=1cm, right= of tmple2, opacity=0.0] (tmple3) {};
      \node[draw, trapezium, minimum width=1cm, fill=yellow!40, rotate around={-90:(tmple3.center)}] (LE3) at (tmple3) {};

      \node[draw, trapezium, minimum width=1cm, right= of tmple3, opacity=0.0] (tmpple4) {};
      \node[draw, trapezium, minimum width=1cm, fill=yellow!40, rotate around={-90:(tmpple4.center)}, opacity=0.0] (PLE4) at (tmpple4) {};

      \draw[->,>=latex] (L) -- node [above, midway] (al1) {1} (LH);
      \draw[->,>=latex] (LH) -- node [above, midway] (al2) {32} (LE1);
      \draw[->,>=latex] (LE1) -- node [above, midway] (al3) {64} (LE2);
      \draw[->,>=latex] (LE2) -- node [above, midway] (al4) {128} (LE3);
      \draw[->,>=latex, opacity=0.0] (LE3) -- node [above, midway] (al5) {} (PLE4);

      \node[draw, minimum width=0.75cm, minimum height=0.75cm, fill=gray!40, below= of al2] (CGRUL1) {};
      \node[draw, minimum width=0.75cm, minimum height=0.75cm, fill=gray!40, below= of al3] (CGRUL2) {};
      \node[draw, minimum width=0.75cm, minimum height=0.75cm, fill=gray!40, below= of al4] (CGRUL3) {};
      \node[draw, minimum width=0.75cm, minimum height=0.75cm, fill=gray!40, below= of al5] (CGRUL4) {};

      \draw[->,>=latex] (al2.south) -- (CGRUL1.north);
      \draw[->,>=latex] (al3.south) -- (CGRUL2.north);
      \draw[->,>=latex] (al4.south) -- (CGRUL3.north);
      \draw[->,>=latex] (LE3.north) -| node [above, midway] {256} (CGRUL4.north);

      \node[below=4.85cm of L] (E) {\includegraphics[width=0.2\textwidth]{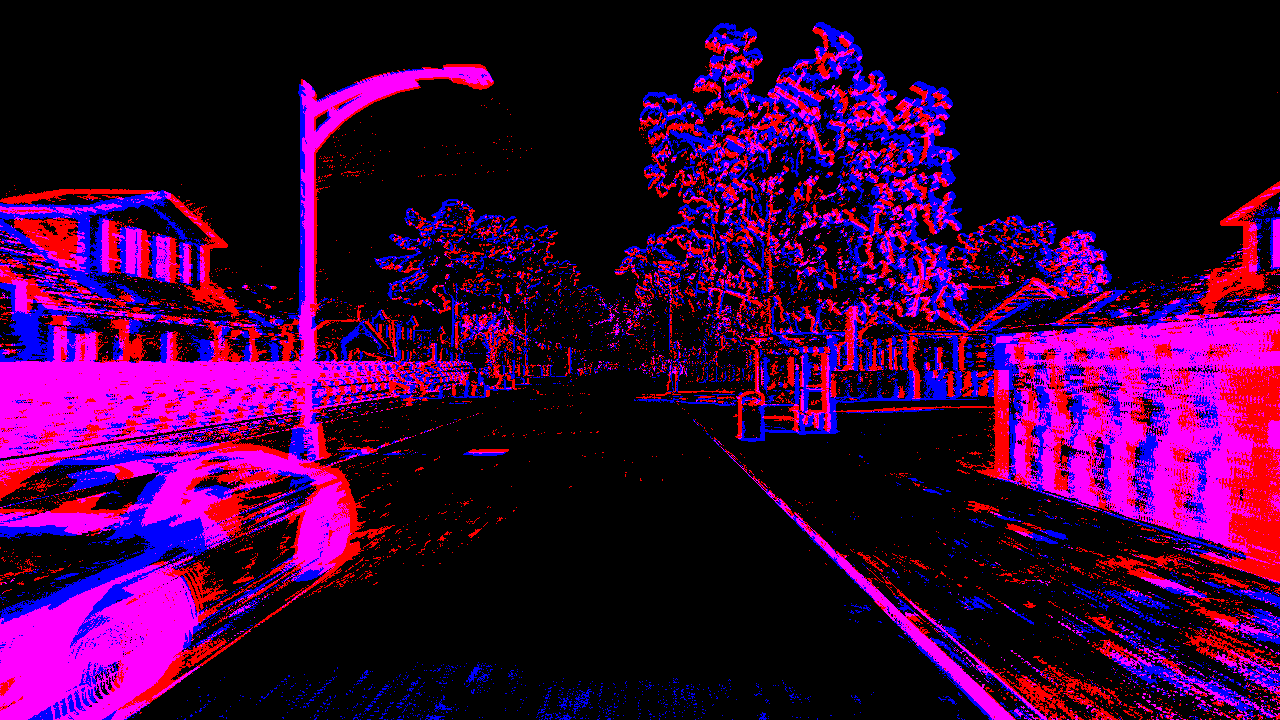}};
      \node[above=0cm of E] (El) {Events};

      \node[draw, trapezium, minimum width=1cm, right= of E, opacity=0.0] (tmpeh) {};
      \node[draw, trapezium, minimum width=1cm, fill=blue!40, rotate around={-90:(tmpeh.center)}] (EH) at (tmpeh) {};

      \node[draw, trapezium, minimum width=1cm, right= of tmpeh, opacity=0.0] (tmpee1) {};
      \node[draw, trapezium, minimum width=1cm, fill=red!40, rotate around={-90:(tmpee1.center)}] (EE1) at (tmpee1) {};

      \node[draw, trapezium, minimum width=1cm, right= of tmpee1, opacity=0.0] (tmpee2) {};
      \node[draw, trapezium, minimum width=1cm, fill=red!40, rotate around={-90:(tmpee2.center)}] (EE2) at (tmpee2) {};

      \node[draw, trapezium, minimum width=1cm, right= of tmpee2, opacity=0.0] (tmpee3) {};
      \node[draw, trapezium, minimum width=1cm, fill=red!40, rotate around={-90:(tmpee3.center)}] (EE3) at (tmpee3) {};

      \node[draw, trapezium, minimum width=1cm, right= of tmpee3, opacity=0.0] (tmppee4) {};
      \node[draw, trapezium, minimum width=1cm, fill=red!40, rotate around={-90:(tmppee4.center)}, opacity=0.0] (PEE4) at (tmppee4) {};

      \draw[->,>=latex] (E) -- node [below, midway] (ae1) {10} (EH);
      \draw[->,>=latex] (EH) -- node [below, midway] (ae2) {32} (EE1);
      \draw[->,>=latex] (EE1) -- node [below, midway] (ae3) {64} (EE2);
      \draw[->,>=latex] (EE2) -- node [below, midway] (ae4) {128} (EE3);
      \draw[->,>=latex, opacity=0.0] (EE3) -- node [below, midway] (ae5) {} (PEE4);

      \node[draw, minimum width=0.75cm, minimum height=0.75cm, fill=gray!40, above= of ae2] (CGRUE1) {};
      \node[draw, minimum width=0.75cm, minimum height=0.75cm, fill=gray!40, above= of ae3] (CGRUE2) {};
      \node[draw, minimum width=0.75cm, minimum height=0.75cm, fill=gray!40, above= of ae4] (CGRUE3) {};
      \node[draw, minimum width=0.75cm, minimum height=0.75cm, fill=gray!40, above= of ae5] (CGRUE4) {};

      \draw[->,>=latex] (ae2.north) -- (CGRUE1.south);
      \draw[->,>=latex] (ae3.north) -- (CGRUE2.south);
      \draw[->,>=latex] (ae4.north) -- (CGRUE3.south);
      \draw[->,>=latex] (EE3.north) -| node [below, midway] {256} (CGRUE4.south);

      \node[draw, minimum width=1cm, minimum height=1cm, below= of CGRUL1] (S1) {1/1};
      \node[draw, minimum width=1cm, minimum height=1cm, below= of CGRUL2] (S2) {1/2};
      \node[draw, minimum width=1cm, minimum height=1cm, below= of CGRUL3] (S3) {1/4};
      \node[draw, minimum width=1cm, minimum height=1cm, below= of CGRUL4] (S4) {1/8};

      \draw[->,>=latex] (CGRUL1.south) -- node [right, midway, align=center] {\textcolor{OliveGreen}{32}\\+\textcolor{Plum}{32}} (S1.north);
      \draw[] (S1.west) -- ($(S1.west)+(-0.25cm,0)$);
      \draw[->,>=latex] ($(S1.west)+(-0.25cm,0)$) |- (CGRUL1.west);
      \draw[->,>=latex] (CGRUL2.south) -- node [right, midway, align=center] {\textcolor{OliveGreen}{64}\\+\textcolor{Plum}{64}} (S2.north);
      \draw[] (S2.west) -- ($(S2.west)+(-0.25cm,0)$);
      \draw[->,>=latex] ($(S2.west)+(-0.25cm,0)$) |- (CGRUL2.west);
      \draw[->,>=latex] (CGRUL3.south) -- node [right, midway, align=center] {\textcolor{OliveGreen}{128}\\+\textcolor{Plum}{128}} (S3.north);
      \draw[] (S3.west) -- ($(S3.west)+(-0.25cm,0)$);
      \draw[->,>=latex] ($(S3.west)+(-0.25cm,0)$) |- (CGRUL3.west);
      \draw[->,>=latex] (CGRUL4.south) -- node [right, midway, align=center] {\textcolor{OliveGreen}{256}} (S4.north);
      \draw[] (S4.west) -- ($(S4.west)+(-0.25cm,0)$);
      \draw[->,>=latex] ($(S4.west)+(-0.25cm,0)$) |- (CGRUL4.west);

      \draw[->,>=latex] (CGRUE1.north) -- node [right, midway, align=center] {\textcolor{OliveGreen}{32}\\+\textcolor{Plum}{32}} (S1.south);
      \draw[->,>=latex] ($(S1.west)+(-0.25cm,0)$) |- (CGRUE1.west);
      \draw[->,>=latex] (CGRUE2.north) -- node [right, midway, align=center] {\textcolor{OliveGreen}{64}\\+\textcolor{Plum}{64}} (S2.south);
      \draw[->,>=latex] ($(S2.west)+(-0.25cm,0)$) |- (CGRUE2.west);
      \draw[->,>=latex] (CGRUE3.north) -- node [right, midway, align=center] {\textcolor{OliveGreen}{128}\\+\textcolor{Plum}{128}} (S3.south);
      \draw[->,>=latex] ($(S3.west)+(-0.25cm,0)$) |- (CGRUE3.west);
      \draw[->,>=latex] (CGRUE4.north) -- node [right, midway, align=center] {\textcolor{OliveGreen}{256}} (S4.south);
      \draw[->,>=latex] ($(S4.west)+(-0.25cm,0)$) |- (CGRUE4.west);

      \node[draw, trapezium, minimum width=1cm, below left=1.5cm and -0.45cm of E, opacity=0.0] (tmpllh) {};
      \node[draw, trapezium, minimum width=1cm, fill=ForestGreen!40, rotate around={-90:(tmpllh.center)}] (LLH) at (tmpllh) {};
      \node[right=0.1cm of tmpllh] (LLHT) {LiDAR head};

      \node[draw, trapezium, minimum width=1cm, below= of tmpllh, opacity=0.0] (tmplle) {};
      \node[draw, trapezium, minimum width=1cm, fill=yellow!40, rotate around={-90:(tmplle.center)}] (LLE) at (tmplle) {};
      \node[right=0.1cm of tmplle] (LLET) {LiDAR encoder};

      \node[draw, trapezium, minimum width=1cm, right= of LLHT, opacity=0.0] (tmpleh) {};
      \node[draw, trapezium, minimum width=1cm, fill=blue!40, rotate around={-90:(tmpleh.center)}] (LEH) at (tmpleh) {};
      \node[right=0.1cm of tmpleh] (LEHT) {Events head};

      \node[draw, trapezium, minimum width=1cm, below= of tmpleh, opacity=0.0] (tmplee) {};
      \node[draw, trapezium, minimum width=1cm, fill=red!40, rotate around={-90:(tmplee.center)}] (LEE) at (tmplee) {};
      \node[right=0.1cm of tmplee] (LEET) {Events encoder};

      \node[draw, minimum width=0.75cm, minimum height=0.75cm, fill=gray!40, right= of LEHT] (LCGRU) {};
      \node[right=0.1cm of LCGRU] (LCGRUT) {convGRU};

      \node[draw, minimum width=1cm, minimum height=1cm, right=0.425cm of LEET] (LS) {\(S\)};
      \node[right=0.1cm of LS, align=left] (LST) {convGRU state\\(scale \(S\))};
    \end{tikzpicture}
  }
  \caption{The encoder part of the network.}\label{fig:encoder}

  \vspace{10mm}

  \resizebox{\textwidth}{!}{
    \begin{tikzpicture}
      \node[draw, minimum width=1cm, minimum height=1cm] (S4) {1/8};
      \node[draw, minimum width=0.75cm, minimum height=0.75cm, fill=gray, below= of S4] (Res1) {};
      \node[draw, minimum width=0.75cm, minimum height=0.75cm, fill=gray, right=0.75cm of Res1] (Res2) {};

      \node[draw, trapezium, minimum width=1cm, right=0.75cm of Res2, opacity=0.0] (tmpu1) {};
      \node[draw, trapezium, minimum width=1cm, fill=orange!40, rotate around={90:(tmpu1.center)}] (U1) at (tmpu1) {};
      \node[draw, circle, minimum width=0.5cm, right=0.75cm of tmpu1] (C1) {C};
      \node[draw, minimum width=1cm, minimum height=1cm, above= of C1] (S3) {1/4};
      \node[draw, trapezium, minimum width=1cm, right=0.75cm of C1, opacity=0.0] (tmpconv1) {};
      \node[draw, trapezium, minimum width=1cm, fill=pink, rotate around={-90:(tmpconv1.center)}] (conv1) at (tmpconv1) {};

      \node[draw, trapezium, minimum width=1cm, right=0.75cm of tmpconv1, opacity=0.0] (tmpu2) {};
      \node[draw, trapezium, minimum width=1cm, fill=orange!40, rotate around={90:(tmpu2.center)}] (U2) at (tmpu2) {};
      \node[draw, circle, minimum width=0.5cm, right=0.5cm of tmpu2] (C2) {C};
      \node[draw, minimum width=1cm, minimum height=1cm, above= of C2] (S2) {1/2};
      \node[draw, trapezium, minimum width=1cm, right=0.75cm of C2, opacity=0.0] (tmpconv2) {};
      \node[draw, trapezium, minimum width=1cm, fill=pink, rotate around={-90:(tmpconv2.center)}] (conv2) at (tmpconv2) {};

      \node[draw, trapezium, minimum width=1cm, right=0.5cm of tmpconv2, opacity=0.0] (tmpu3) {};
      \node[draw, trapezium, minimum width=1cm, fill=orange!40, rotate around={90:(tmpu3.center)}] (U3) at (tmpu3) {};
      \node[draw, circle, minimum width=0.5cm, right=0.5cm of tmpu3] (C3) {C};
      \node[draw, minimum width=1cm, minimum height=1cm, above= of C3] (S1) {1/1};
      \node[draw, trapezium, minimum width=1cm, right=0.5cm of C3, opacity=0.0] (tmpconv3) {};
      \node[draw, trapezium, minimum width=1cm, fill=pink, rotate around={-90:(tmpconv3.center)}] (conv3) at (tmpconv3) {};

      \node[draw, trapezium, minimum width=1cm, right=0.5cm of tmpconv3, opacity=0.0] (tmpp) {};
      \node[draw, trapezium, minimum width=1cm, fill=cyan!60, rotate around={-90:(tmpp.center)}] (P) at (tmpp) {};
      \node[right=0.5cm of tmpp] (out) {\includegraphics[width=0.2\textwidth]{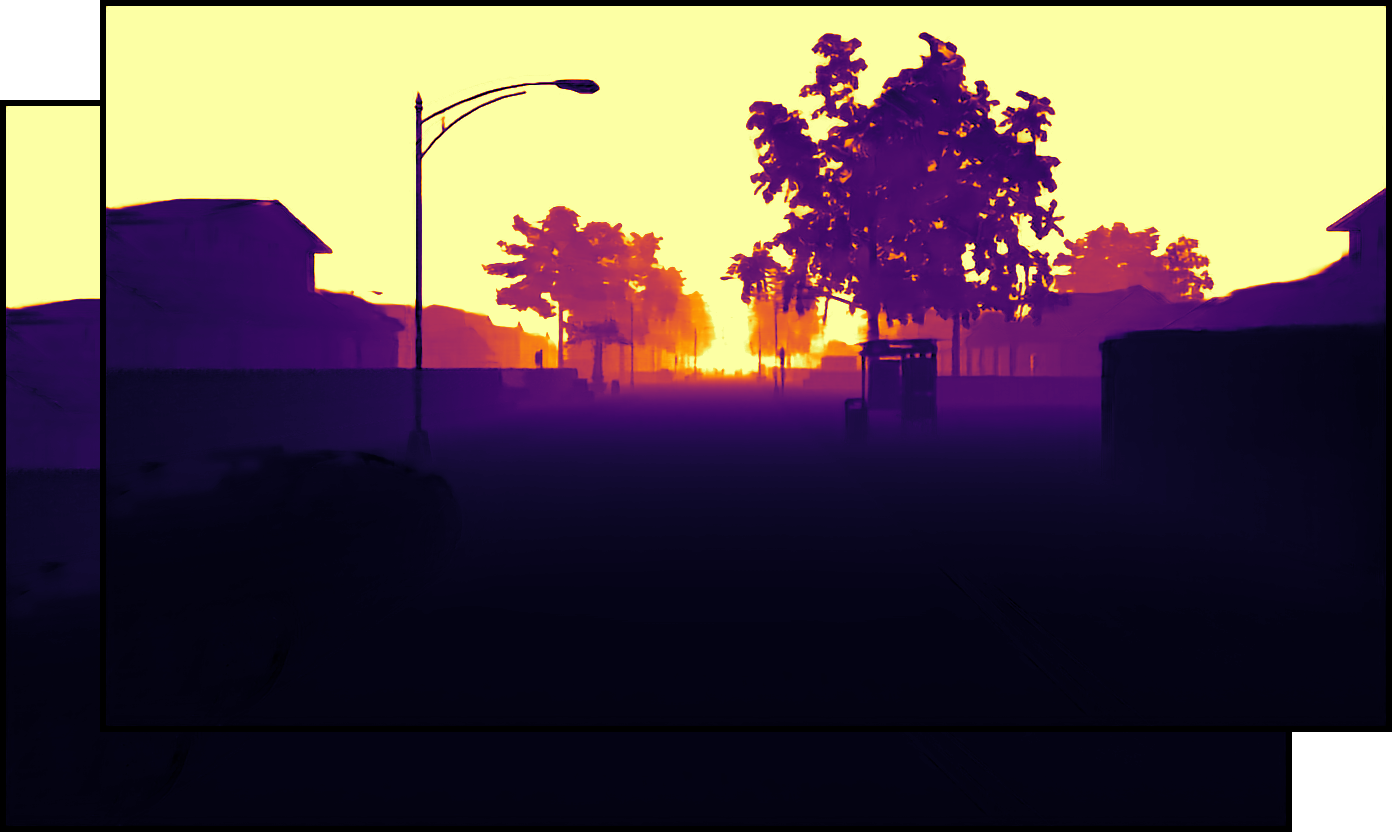}};
      \node[above=0cm of out, align=center] (outl) {Dense depths\\\(D_\text{bf}\) and \(D_\text{af}\)};

      \draw[->,>=latex] (S4) -- node [right, midway, align=center] {\textcolor{OliveGreen}{256}} (Res1);
      \draw[->,>=latex] (Res1) -- node [below, midway, align=center] {256} (Res2);
      \draw[->,>=latex] (Res2) -- node [below, midway, align=center] {256} (U1);
      \draw[->,>=latex] (S3) -| node [above, midway, align=center] {\textcolor{Plum}{128}} (U1);
      \draw[->,>=latex] (S3) -- node [right, midway, align=center] {\textcolor{OliveGreen}{128}} (C1);
      \draw[->,>=latex] (U1) -- node [below, midway, align=center] {128} (C1);
      \draw[->,>=latex] (C1) -- node [below, midway, align=center] {256} (conv1);
      \draw[->,>=latex] (conv1) -- node [below, midway, align=center] {128} (U2);
      \draw[->,>=latex] (S2) -| node [above, midway, align=center] {\textcolor{Plum}{64}} (U2);
      \draw[->,>=latex] (S2) -- node [right, midway, align=center] {\textcolor{OliveGreen}{64}} (C2);
      \draw[->,>=latex] (U2) -- node [below, midway, align=center] {64} (C2);
      \draw[->,>=latex] (C2) -- node [below, midway, align=center] {128} (conv2);
      \draw[->,>=latex] (conv2) -- node [below, midway, align=center] {64} (U3);
      \draw[->,>=latex] (S1) -| node [above, midway, align=center] {\textcolor{Plum}{32}} (U3);
      \draw[->,>=latex] (S1) -- node [right, midway, align=center] {\textcolor{OliveGreen}{32}} (C3);
      \draw[->,>=latex] (U3) -- node [below, midway, align=center] {32} (C3);
      \draw[->,>=latex] (C3) -- node [below, midway, align=center] {64} (conv3);
      \draw[->,>=latex] (conv3) -- node [below, midway, align=center] {32} (P);
      \draw[->,>=latex] (P) -- node [below, midway, align=center] {2} (out);

      \node[draw, minimum width=1cm, minimum height=1cm, below right=1.5cm and 2.5cm of Res1] (LS) {\(S\)};
      \node[right=0.1cm of LS, align=left] (LST) {convGRU state\\(scale \(S\))};

      \node[draw, minimum width=0.75cm, minimum height=0.75cm, fill=gray, right= of LST] (LRes) {};
      \node[right=0.1cm of LRes] (LResT) {Residual block};

      \node[draw, trapezium, minimum width=1cm, right=0.5cm of LResT, opacity=0.0] (tmplu) {};
      \node[draw, trapezium, minimum width=1cm, fill=orange!40, rotate around={90:(tmplu.center)}] (LU) at (tmplu) {};
      \node[right=0.1cm of tmplu] (LUT) {Convex upsampling};

      \node[draw, circle, minimum width=0.5cm, below=0.5cm of LS] (LC) {C};
      \node[right=0.1cm of LC] (LCT) {Concatenation};

      \node[draw, trapezium, minimum width=1cm, below=0.725cm of LRes, opacity=0.0] (tmplconv) {};
      \node[draw, trapezium, minimum width=1cm, fill=pink, rotate around={-90:(tmplconv.center)}] (Lconv) at (tmplconv) {};
      \node[right=0.1cm of tmplconv] (LconvT) {Convolution};

      \node[draw, trapezium, minimum width=1cm, below=0.85cm of tmplu, opacity=0.0] (tmplpred) {};
      \node[draw, trapezium, minimum width=1cm, fill=cyan!60, rotate around={-90:(tmplpred.center)}] (Lpred) at (tmplpred) {};
      \node[right=0.1cm of tmplpred] (LpredT) {Prediction head};
    \end{tikzpicture}
  }
  \caption{The decoder part of the network.}\label{fig:decoder}
\end{figure}

Inspired by the Recurrent Asynchronous Multimodal Network (RAMNet) architecture of Gehrig et al.~\cite{Gehrig2021CombiningEA} for RGB and events fusion, we propose here a fully convolutional recurrent network to estimate dense depths from asynchronous LiDAR and events data. We call it the ALED Net, for Asynchronous LiDAR and Events Depths densification network.

Our network can be decomposed in two main parts: an encoder, tasked with fusing asynchronous events and LiDAR features at different scales, and a decoder, tasked with interpreting the fused features for estimating dense depths.

In its encoder part, illustrated in Fig.~\ref{fig:encoder}, the LiDAR and events inputs are fed independently. Both of them go through an encoding head, computing a first feature map of 32 channels while keeping their original height and width. Convolutional encoders (in the form of ResNet Basic Blocks~\cite{He2016DeepRL}) are then used to compute feature maps at scales \(1/2\), \(1/4\), and \(1/8\), doubling the number of channels every time. Each of these feature maps is then used as the input of convolutional gated recurrent unit (convGRU) blocks~\cite{Siam2017ConvolutionalGR}, updating its corresponding state. Since these states are shared between the LiDAR and events encoders, both parts of the network can update them asynchronously.

In its decoder part, illustrated in Fig.~\ref{fig:decoder}, the convGRU state at the lowest scale first goes through two residual blocks. Then, for each following scale, the decoded feature map from the previous scale is upscaled by using convex upsampling~\cite{Teed2020RAFTRA}. While a simple bilinear upsampling was used in RAMNet~\cite{Gehrig2021CombiningEA}, convex upsampling allows our network to learn how to upscale features from a lower scale, using information from a higher scale. We propose to design the convGRU such that the first half of its state (in purple in Fig.~\ref{fig:encoder}~and~\ref{fig:decoder}) guides the convex upsampling. Fusion of the upsampled decoded features and the state from the current scale is then performed by concatenating the output of the convex upsampling block with the remaining half of the convGRU state (in green in Fig.~\ref{fig:encoder}~and~\ref{fig:decoder}), and by applying a convolution to reduce the number of channels. After the last scale, a prediction head is used to obtain the two final depth maps, \(D_\text{bf}\) and \(D_\text{af}\), in the form of a two-channel tensor, at the same full resolution as the events input.

Regarding the implementation, both encoding heads use a kernel size of 5. LiDAR and events encoders use a kernel size of 5, with stride 2. Both the convGRU and residual blocks use a kernel size of 3. The convolutions in the convex upsampling blocks use a kernel size of 5, while the convolutions following the concatenations use a kernel size of 1. Finally, the prediction layer also uses a kernel size of 1. Convolutions are followed by a PReLU activation function~\cite{He2015DelvingDI}, and instance normalization is used in the ResNet encoders as proposed by Pan et al.~\cite{Pan2018TwoAO}. In total, the network contains 26 millions of trainable parameters.

\subsection{Data Representation}

\subsubsection{Events}
We use Discretized Event Volumes~\cite{Zhu2019UnsupervisedEL} as the input representation for the events. We follow the formulation of Perot et al.~\cite{Perot2020LearningTD}, where the Discretized Event Volume \(V\) for input events \(\{e_i = (x_i, y_i, p_i, t_i)\}_{i=1}^N\) is described as:
\begin{equation}
V_{t, p, y, x} = \sum_{e_i, x_i=x, y_i=y, p_i=p} \max(0, 1-|t-t_i^*|)
\end{equation}
\begin{equation}
t_i^* = (B-1)\frac{t_i-t_0}{t_N-t_0}
\end{equation}
where \((x, y)\) is the position of the event, \(t\) its timestamp, and \(p\) its polarity.

In our experiments, we set \(B=5\text{ bins}\), and concatenate the negative and positive polarity bins along the first dimension, resulting in a tensor of shape \((10, H, W)\).

\subsubsection{LiDAR and Depths}
LiDAR data is fed to the network as a 1-channel depth image. To do so, each LiDAR point cloud is projected onto the image plane of the event camera. Pixels where one or more LiDAR points fall into are given as value the lowest depth. Pixels without any LiDAR point are given a value of 0. For an easier learning, the LiDAR projection and ground truth images are normalized between 0 and 1 based on the maximum LiDAR range (200m in the case of our synthetic SLED dataset, 100m in the case of the MVSEC dataset~\cite{Zhu2018TheMS}).

\subsection{Loss functions}
To train our network, we combine the use of two losses: a pixel-wise \(\ell_1\) loss \(\mathcal{L}_\text{pw}\), and a multiscale gradient matching loss \(\mathcal{L}_\text{msg}\).

The pixel-wise \(\ell_1\) loss operates as the main supervision loss, applied on both the ``before'' and the ``after'' depth maps, and is defined as follows:
\begin{equation}
  \mathcal{L}_\text{pw} = \sum_{x, y} \left\lVert D(x, y) - \hat{D}(x, y) \right\rVert _1
\end{equation}
where \(D\) and \(\hat{D}\) are the respectively the estimated and ground truth depth maps. 

However, when supervised by the \(\ell_1\) loss alone, the network tends to produce blurry and non-smooth depth images. To solve this issue, we use here a multiscale gradient matching loss inspired by~\cite{Ummenhofer2017DeMoNDA}, also applied on both the ``before'' and the ``after'' depth maps, and defined as
\begin{equation}
  \mathcal{L}_\text{msg} = \sum_{h \in \{1, 2, 4, 8, 16\}} \sum_{x, y} \left\lVert \text{\textbf{g}}[D](x, y, h) - \text{\textbf{g}}[\hat{D}](x, y, h) \right\rVert _2
\end{equation}
with the discrete gradient function \textbf{g} of an image \(f\) defined as
\begin{equation}
  \text{\textbf{g}}[f](x, y, h) = \left( f(x+h, y) - f(x, y); f(x, y+h) - f(x, y) \right) ^T
\end{equation}
This loss helps regulating the depth results, by making depth discontinuities more prominent, and by smoothing homogeneous regions.


Our total loss \(\mathcal{L}\) for a sequence of length \(T\) is finally defined as
\begin{equation}
  \mathcal{L} = \sum_{t=0}^T \sum_{\text{bf}, \text{af}} (\mathcal{L}_{\text{pw}}^t + \alpha \mathcal{L}_{\text{msg}}^t)
\end{equation}
where \(\alpha\) is a weight parameter for the multiscale gradient matching loss.

In our experiments, we observed that giving too much importance to the multiscale gradient matching loss early in the training makes the network unable to derive correct depth estimates. Therefore, we always set \(\alpha = 0.1\) during the first epoch of training, to force the network to use mainly the \(\ell_1\) loss and converge towards good initial depth estimates. For the remaining epochs, we set \(\alpha = 1\).

\section{The SLED Dataset}
In order to train and evaluate the proposed network, we require a dataset containing both events, LiDAR point clouds, as well as a dense ground truth on depths. While we can use the MVSEC dataset~\cite{Zhu2018TheMS} for low-resolution cameras, its ground truth is constructed by accumulating point clouds from a LiDAR sensor, a solution which introduces errors in case of moving objects. A similar dataset does not exist for sensors of higher resolution.

For these reasons, we use the CARLA simulator~\cite{Dosovitskiy2017CARLAAO} (version 0.9.14) to generate a dataset with perfect synchronization and calibration of the sensors, and perfect ground truth depth. We call it SLED, for Synthetic LiDAR Events Depths dataset. It is composed of 160 sequences of 10 seconds each, for a total of more than 20 minutes of data. These sequences are recorded on the \textit{Town01} to \textit{Town07} and \textit{Town10HD} maps (20 sequences for each map), each sequence starting from a different geographic location. By doing so, a wide range of environments is represented within the dataset, as detailed in Table~\ref{tab:sled_content}.

\begin{table}
  \centering
  \setlength\tabcolsep{6pt}
  \caption{Detailed content of our SLED dataset containing 160 sequences of 10 seconds each.}\label{tab:sled_content}
  \resizebox{\linewidth}{!}{
    \begin{tabular}{c c c c c c}
      \toprule
      Map & Set & Environment & Features & Night seq. & Day seq. \\
      \midrule
      Town01 & Test & Town & Small buildings, bridges, distant forests and mountains, palm trees & 4 & 16 \\
      Town02 & Train & Town & Small buildings, plazas, forest road & 4 & 16 \\
      Town03 & Test & City & Tall and small buildings, roundabouts, tunnel, aerial railway & 4 & 16 \\
      Town04 & Val. & Town & Small buildings, highway, parking, lake, forests and mountains & 4 & 16 \\
      Town05 & Train & City & Tall buildings, parking, aerial beltway and railway & 4 & 16 \\
      Town06 & Train & Suburban & Small buildings, U-turns, distant hills & 4 & 16 \\
      Town07 & Train & Countryside & Barns, grain silos, fields, mountain road & 4 & 16 \\
      Town10HD & Train & City & Buildings, monuments and sculptures, playgrounds, seaside & 4 & 16 \\
      \bottomrule
    \end{tabular}
  }
\end{table}

Each sequence contains a 1280\texttimes{}720 event camera, a 40-channel LiDAR, and a 1280\texttimes{}720 depth camera which is perfectly aligned with the event camera. Both the event data and the depth images are recorded at 200Hz, while the LiDAR is configured to run at 10Hz. RGB images (1280\texttimes{}720) are also provided at a 30Hz rate, aligned with the event-based sensor. The LiDAR sensor is configured with a maximum range of 200 meters. For realism and diversity purposes, AI-controlled vehicles and pedestrians are added to the simulation. Sun altitude also varies, resulting for each map in 4 night recordings, and the other 16 recordings ranging from early morning (where the sun can be directly in front of the camera) to midday (where the sun is at its apogee). Varying cloudiness conditions are also used, adding more or less texture to the sky, and making shadows more diverse.

We also configure the event camera in CARLA to use a linear intensity scale rather than the default logarithmic one, making the events produced by the simulator more realistic. More details on that topic, as well as an overview of the data contained in the dataset, are given in the Supplementary Material.

\section{Evaluation}

\subsection{Dense Depths}

\subsubsection{On the SLED Dataset}
For training on the SLED dataset, we use the Adam optimizer~\cite{Kingma2015AdamAM} with a learning rate of \(10^{-4}\) and a batch size of 4, and train for a total of 50 epochs. To augment input data, we randomly crop it to \(608 \times 608\), and apply random horizontal flipping.

\begin{table}
  \centering
  \setlength\tabcolsep{6pt}
  \caption{Errors on the testing set of the SLED dataset for various cutoff depth distances. From left to right: average absolute and relative depth errors on both the ``before'' \(D_\text{bf}\) and ``after'' \(D_\text{af}\) depth maps; average absolute depth errors when associating a depth to each event; average absolute depth difference errors and percentage of correctly classified events based on this depth difference.}\label{tab:results_sled}
  \resizebox{\linewidth}{!}{
    \begin{tabular}{c c cccc cccc cc}
      \toprule
      \multirow{3}{*}{\textbf{Map}} & \multirow{3}{*}{\textbf{Cutoff}} & \multicolumn{4}{c}{\textbf{Dense depths errors}} & \multicolumn{4}{c}{\textbf{Sparse depths errors}} & \multicolumn{2}{c}{\textbf{Depth change map errors}} \\
      & & \multicolumn{2}{c}{On \(D_\text{bf}\)} & \multicolumn{2}{c}{On \(D_\text{af}\)} & \multicolumn{2}{c}{On \(D_\text{bf}\)} & \multicolumn{2}{c}{On \(D_\text{af}\)} & \multirow{2}{*}{Absolute error} & Correctly classified events \\
      & & Raw & Rel. & Raw & Rel. & NN & ALED\textsubscript{S} & NN & ALED\textsubscript{S} & & (with a threshold of \rpm{}1m) \\
      \midrule
      \multirow{5}{*}{Town01} & 10m & 1.24m & 20.99\% & 1.37m & 23.60\% & 1.32m & 1.46m & 2.24m & 1.79m & 2.11m & 90.27\% \\
      & 20m & 2.08m & 23.06\% & 2.27m & 25.48\% & 1.51m & 1.84m & 2.53m & 2.15m & 3.18m & 85.07\% \\
      & 30m & 2.72m & 23.76\% & 2.92m & 26.03\% & 1.71m & 2.37m & 2.83m & 2.67m & 3.88m & 81.68\% \\
      & 100m & 4.25m & 24.01\% & 4.51m & 26.07\% & 2.40m & 3.48m & 3.91m & 3.95m & 5.12m & 77.48\% \\
      & 200m & 4.53m & 17.20\% & 4.81m & 18.66\% & 7.86m & 5.44m & 9.76m & 6.23m & 7.36m & 75.54\% \\
      \midrule
      \multirow{5}{*}{Town03} & 10m & 2.00m & 28.91\% & 2.09m & 30.11\% & 0.47m & 0.56m & 0.67m & 0.66m & 1.14m & 93.70\% \\
      & 20m & 2.85m & 29.91\% & 2.97m & 31.15\% & 0.64m & 0.75m & 1.12m & 0.87m & 2.54m & 87.16\% \\
      & 30m & 3.33m & 29.10\% & 3.45m & 30.24\% & 0.92m & 1.11m & 1.61m & 1.26m & 3.23m & 83.71\% \\
      & 100m & 4.60m & 27.37\% & 4.77m & 28.42\% & 1.88m & 2.55m & 3.17m & 2.88m & 4.47m & 78.50\% \\
      & 200m & 4.86m & 21.50\% & 5.03m & 22.33\% & 4.43m & 3.60m & 5.93m & 4.10m & 6.20m & 77.23\% \\
      \bottomrule
    \end{tabular}
  }
\end{table}

\begin{figure}
  \centering
  \includegraphics[width=0.35\linewidth]{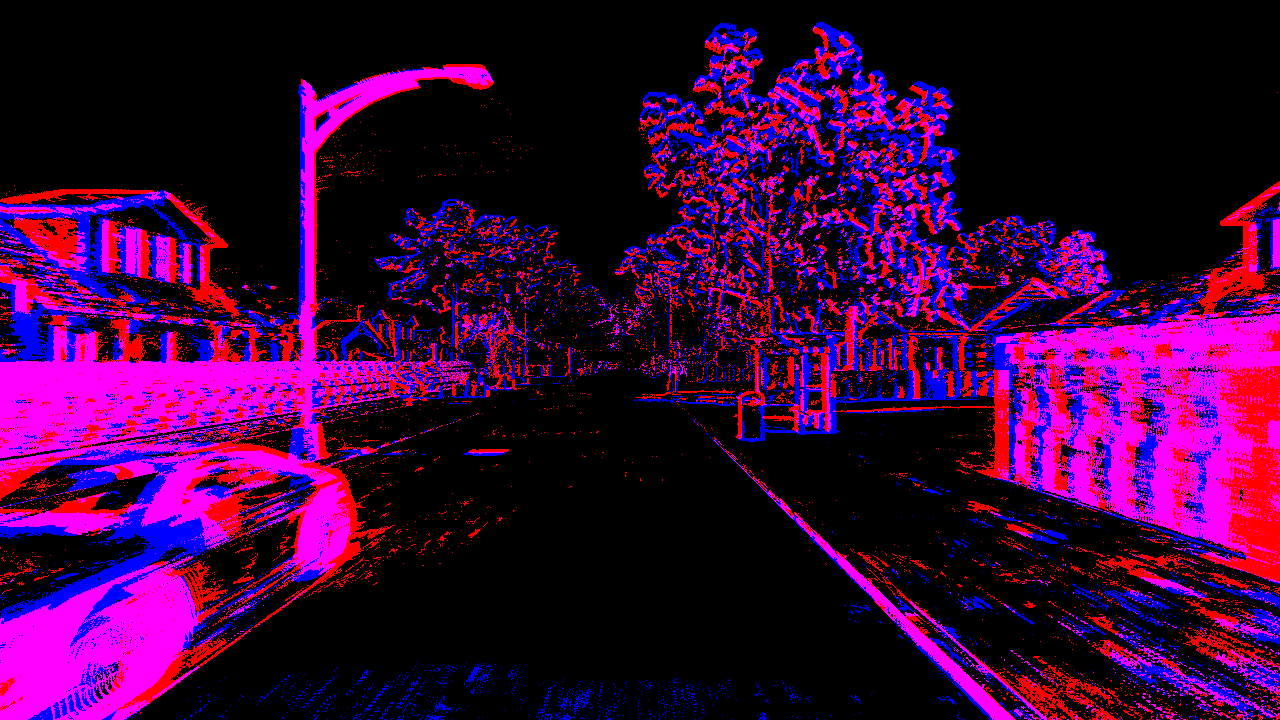}
  \includegraphics[width=0.35\linewidth]{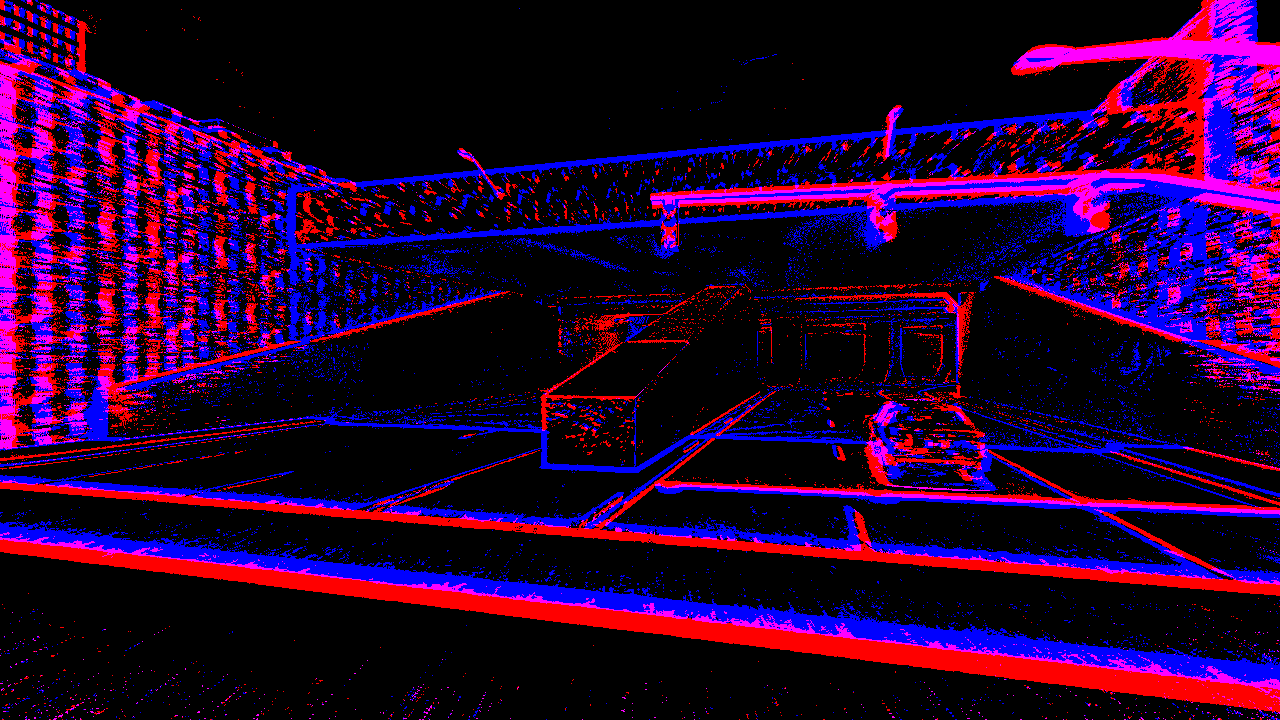} \\
  \includegraphics[width=0.35\linewidth]{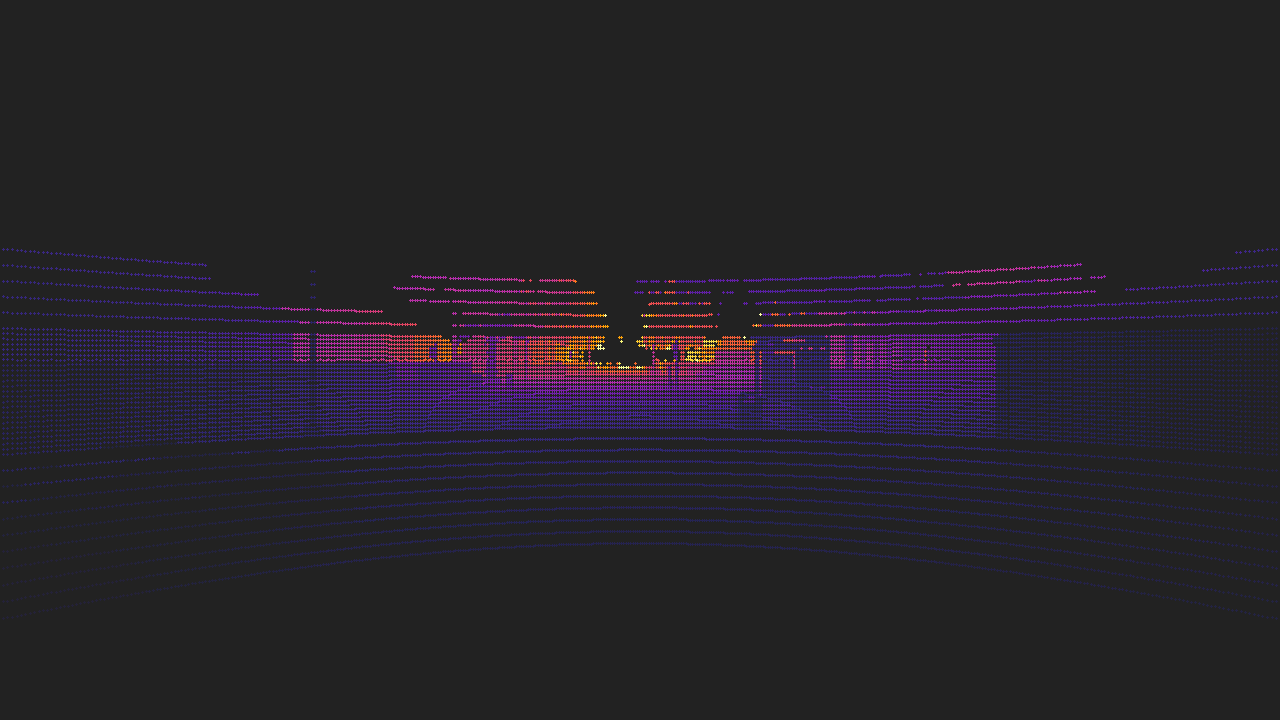}
  \includegraphics[width=0.35\linewidth]{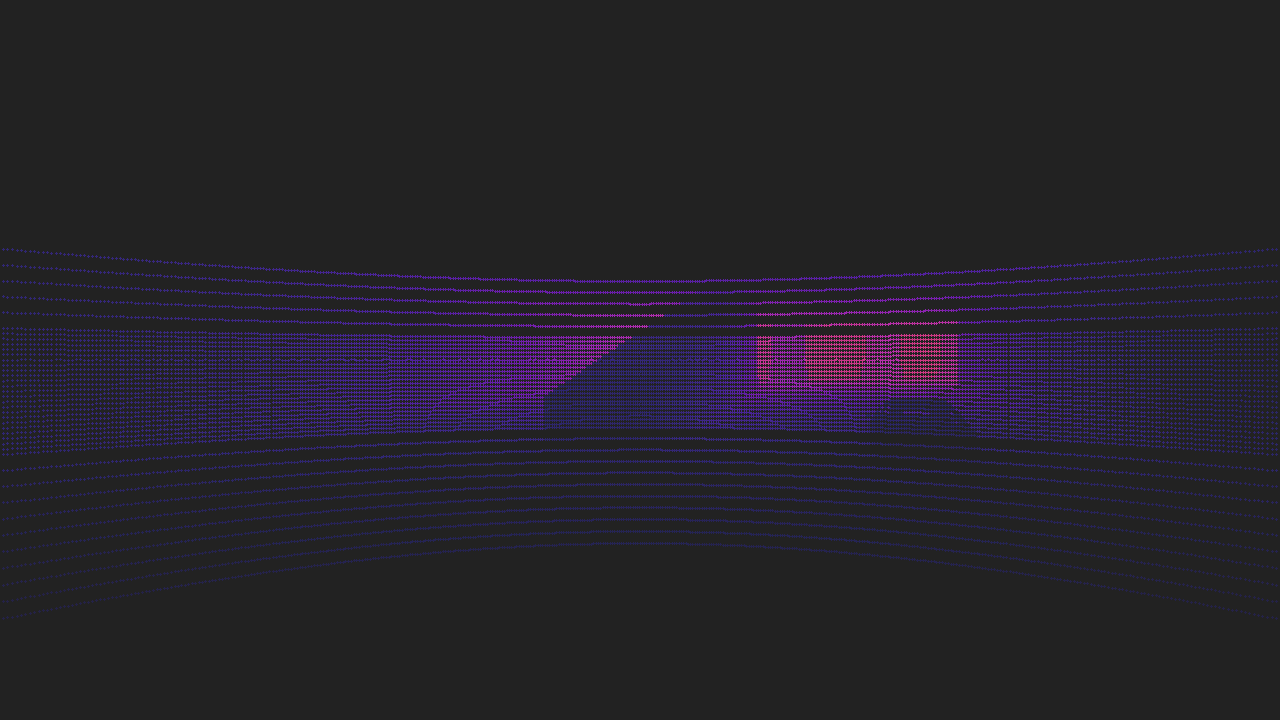} \\
  \includegraphics[width=0.35\linewidth]{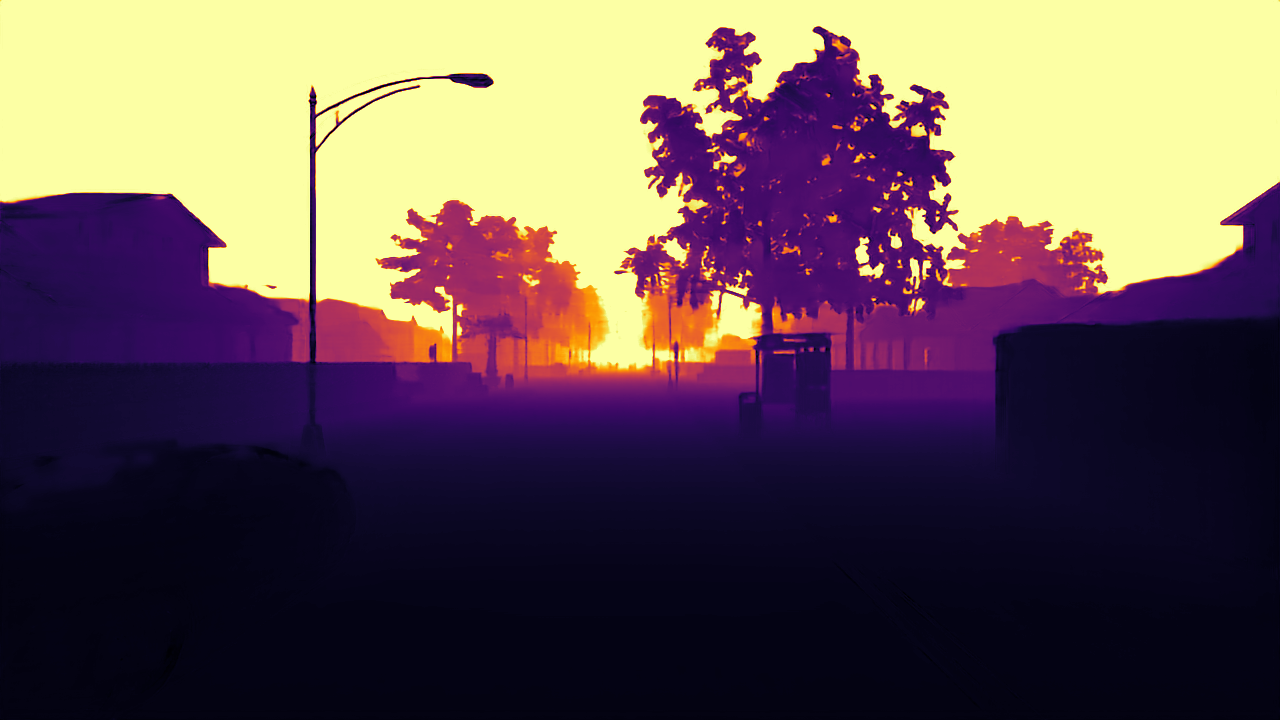}
  \includegraphics[width=0.35\linewidth]{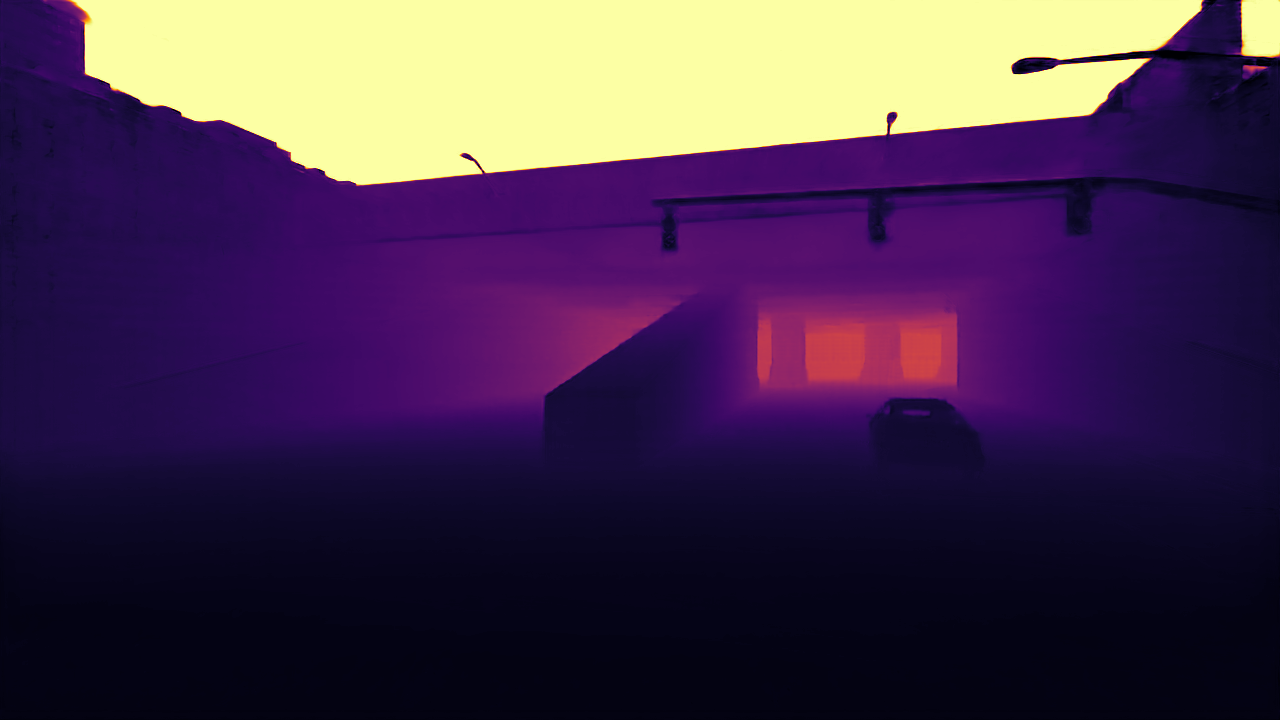} \\
  \includegraphics[width=0.35\linewidth]{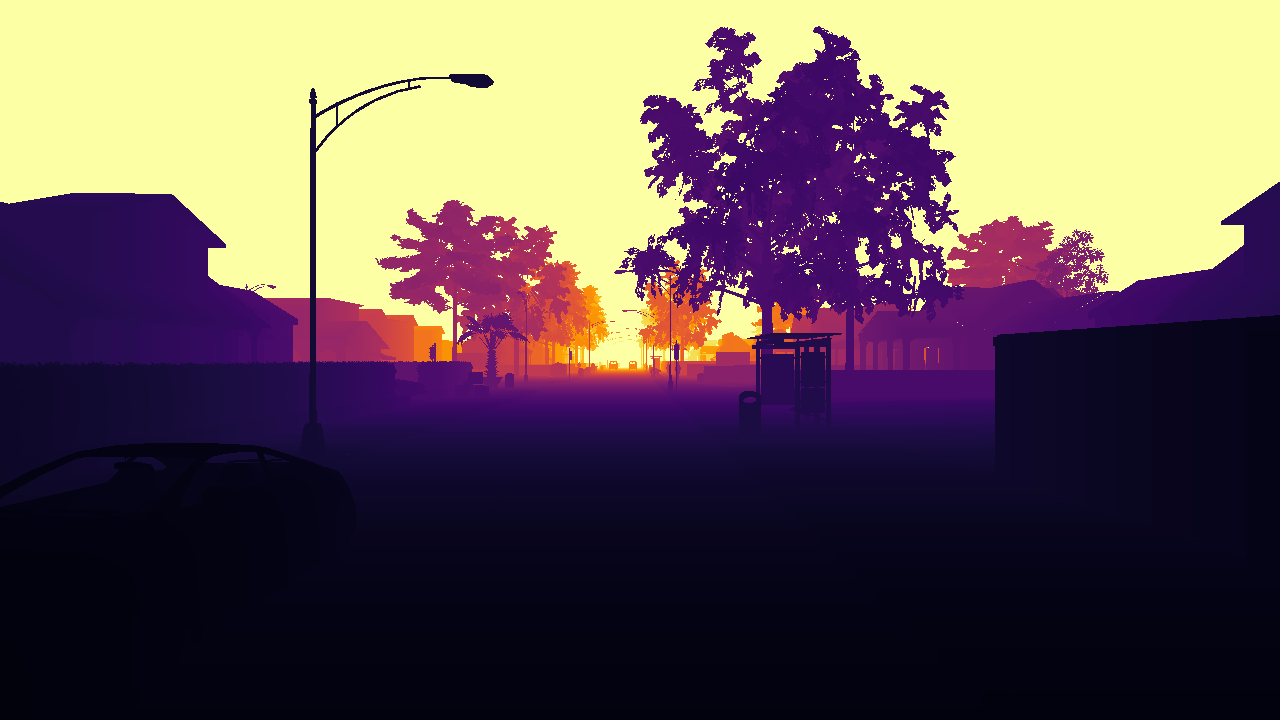}
  \includegraphics[width=0.35\linewidth]{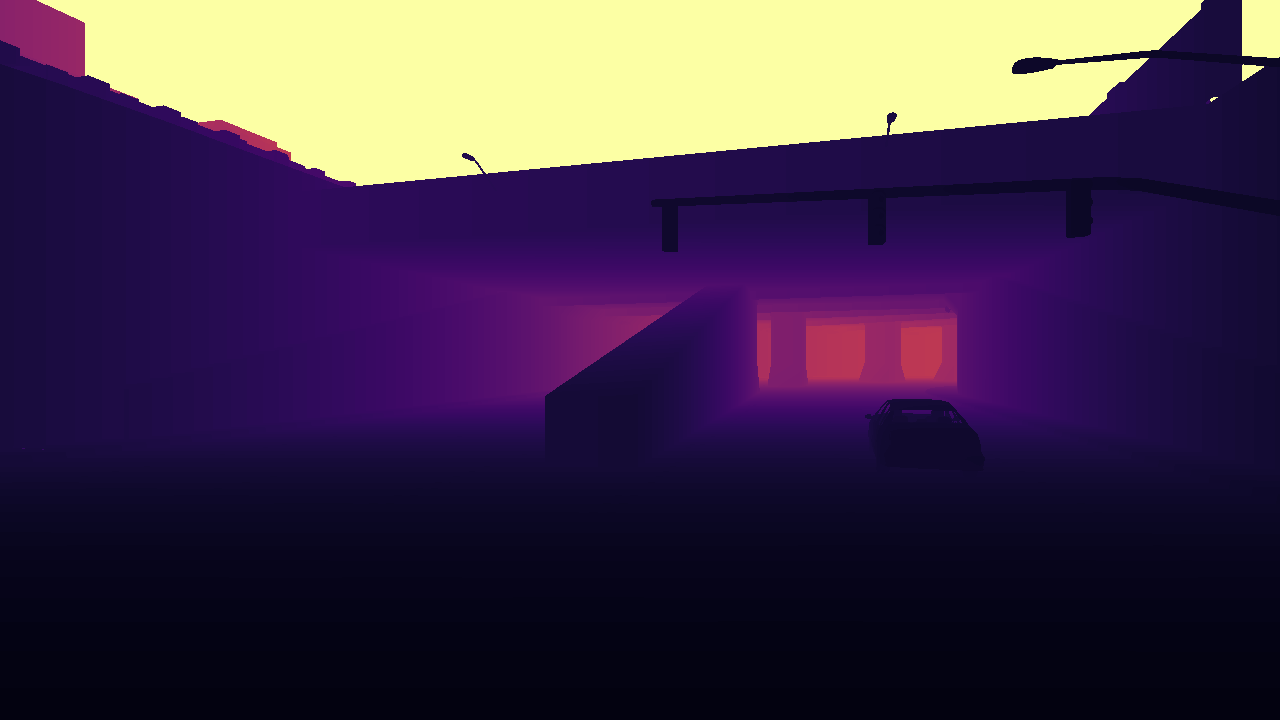} \\
  \includegraphics[width=0.35\linewidth]{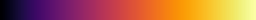} \\
  \begin{tikzpicture}
    \node[] (black) {0m};
    \node[right=1.1cm of black] (pink) {100m};
    \node[right=0.8cm of pink] (yellow) {200m};
  \end{tikzpicture}
  \caption{Two qualitative results on our synthetic SLED dataset, on \textit{Town01} (left) and \textit{Town02} (right). From top to bottom: events, LiDAR, predicted depth map, ground truth, color scale.}\label{fig:sled_dense_cmp}
\end{figure}
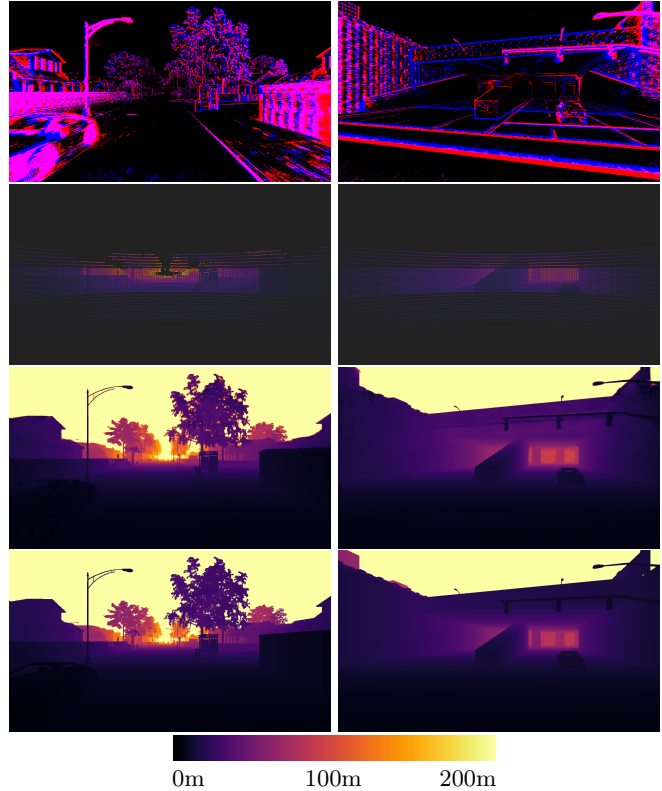

Numerical results on the testing set are presented in the ``Dense depths errors'' column of Table~\ref{tab:results_sled}. Evaluations are conducted on \textit{Town01} and \textit{Town03} maps, which contain challenging environments with many unique features (bridges, tunnels, \dots) that are not present in the training maps. For the max range of 200 meters, ALED estimates depth maps with an average absolute error slightly over 4.5 meters for \textit{Town01}, and around 5 meters for \textit{Town03}. The respective average relative error is around 18\% for \textit{Town01}, and around 22\% for \textit{Town03}. A first element that can explain these errors is that the LiDAR has a small vertical coverage of the image: close ground objects or the top of close buildings are not reached by the LiDAR, meaning that accurate depth estimations for these objects are complex to achieve. In opposition, sky pixels (for which ALED produces good results) are only accounted for at the full 200m cutoff. These observations can be correlated to the larger relative errors observed for close cutoff distances than for the full 200m cutoff. It can also be observed that errors on the ``after'' depth maps are slightly higher. This is to be expected, as the network has to make use of the events to estimate the movement and propagate the depths accordingly. If the reader is interested, results for each sequence of the testing set are given in the Supplementary Material.

Qualitative results are given in Fig.~\ref{fig:sled_dense_cmp}. They showcase the ability of the network to estimate accurate depths for the whole image, by using events as a guide for the areas the LiDAR sensor can not reach. This is particularly visible for the trees and the light pole for the left column, or the ceiling of the tunnel for the right column. If the reader is interested, more visual results are given in the Supplementary Material and in the videos linked in the introduction.

\subsubsection{On the MVSEC Dataset}

In order to be able to compare our results with the other approaches in the literature, we also train and evaluate ALED on the MVSEC dataset~\cite{Zhu2018TheMS}. We conduct our evaluation under three different sets of weights from different training setups described below:
\begin{itemize}
  \item ALED\textsubscript{S}: the network is only trained in simulation, on proposed SLED dataset;
  \item ALED\textsubscript{R}: the network is only trained on real data, on the MVSEC dataset;
  \item ALED\textsubscript{S\(\rightarrow\)R}: the network is first trained on the SLED dataset, then fined-tuned on the MVSEC dataset.
\end{itemize}

We use a batch size of 4 and a learning rate of \(10^{-4}\) (\(10^{-5}\) when fine-tuning). 50 epochs are used when training the network from zero, 5 when fine-tuning it. We also augment input data, by randomly cropping it to \(256 \times 256\), and by applying random horizontal flipping. Numerical results are given in Table~\ref{tab:results_mvsec}.

Comparing our three sets of trained weights, it appears clearly that training on synthetic data before fine-tuning the network on the MVSEC dataset (\(S\rightarrow R\)) produces the best results. Training from zero on the MVSEC dataset (\(R\)) is not as good as the \(S\rightarrow R\) variant due to the limited data available for training. Finally, training solely on the SLED dataset (\(S\)) produces the worst results, due to the large differences in terms of resolution and LiDAR models between the two datasets, and as simulation is not a perfect reproduction of real data.

\begin{table}
  \setlength\tabcolsep{6pt}
  \caption{Average absolute depth errors (in meters) on the MVSEC dataset for various cutoff depth distances. This evaluation is performed on the ``before'' depth map \(D_\text{bf}\), to be consistent with the methods we compare ourselves to.}\label{tab:results_mvsec}
  \resizebox{\linewidth}{!}{
    \begin{tabular}{c c cc cc cccc}
    \toprule
    \multirow{2}{*}{\textbf{Recording}} & \multirow{2}{*}{\textbf{Cutoff}} & \multicolumn{2}{c}{\textbf{Event-based}} & \multicolumn{2}{c}{\textbf{Event- and frame-based}} & \multicolumn{4}{c}{\textbf{LiDAR- and event-based}} \\
    & & Zhu et al.~\cite{Zhu2019UnsupervisedEL} & E2Depth~\cite{HidalgoCarrio2020LearningMD} & RAMNet~\cite{Gehrig2021CombiningEA} & EvT\textsuperscript{+}~\cite{Sabater2022EventTA} & Cui et al.~\cite{Cui2022DenseDE} & \textbf{ALED\textsubscript{S}} & \textbf{ALED\textsubscript{R}} & \textbf{ALED\textsubscript{S\(\rightarrow\)R}} \\
    \midrule
    \multirow{5}{*}{Outdoor day 1} & 10m & 2.72 & 1.85 & 1.39 & 1.27 & 1.24 & 1.54 & \underline{0.91} & \textbf{0.50} \\
    & 20m & 3.84 & 2.64 & 2.17 & 1.94 & 1.28 & 2.55 & \underline{1.22} & \textbf{0.80} \\
    & 30m & 4.40 & 3.13 & 2.76 & 2.37 & 4.87 & 3.18 & \underline{1.43} & \textbf{1.02} \\
    & 50m & - & - & - & - & - & 3.79 & \underline{1.67} & \textbf{1.31} \\
    & 100m & - & - & - & - & - & 4.08 & \underline{1.96} & \textbf{1.60} \\
    \midrule
    \multirow{5}{*}{Outdoor night 1} & 10m & 3.13 & 3.38 & 2.50 & \textbf{1.48} & 2.26 & 2.24 & 1.75 & \underline{1.52} \\
    & 20m & 4.02 & 3.82 & 3.19 & \underline{2.09} & 2.19 & 3.32 & 2.10 & \textbf{1.81} \\
    & 30m & 4.89 & 4.46 & 3.82 & 2.84 & 4.50 & 3.82 & \underline{2.25} & \textbf{1.95} \\
    & 50m & - & - & - & - & - & 4.31 & \underline{2.44} & \textbf{2.20} \\
    & 100m & - & - & - & - & - & 4.62 & \underline{2.73} & \textbf{2.54} \\
    \midrule
    \multirow{5}{*}{Outdoor night 2} & 10m & 2.19 & 1.67 & 1.21 & 1.48 & 1.88 & 1.94 & \underline{1.19} & \textbf{1.09} \\
    & 20m & 3.15 & 2.63 & 2.31 & 2.13 & 2.14 & 2.82 & \underline{1.65} & \textbf{1.49} \\
    & 30m & 3.92 & 3.58 & 3.28 & 2.92 & 4.67 & 3.22 & \underline{1.81} & \textbf{1.64} \\
    & 50m & - & - & - & - & - & 3.58 & \underline{1.95} & \textbf{1.80} \\
    & 100m & - & - & - & - & - & 3.78 & \underline{2.11} & \textbf{1.97} \\
    \midrule
    \multirow{5}{*}{Outdoor night 3} & 10m & 2.86 & 1.42 & 1.01 & 1.40 & 1.78 & 1.76 & \underline{0.85} & \textbf{0.81} \\
    & 20m & 4.46 & 2.33 & 2.34 & 2.05 & 1.93 & 2.43 & \underline{1.25} & \textbf{1.16} \\
    & 30m & 5.05 & 3.18 & 3.43 & 2.79 & 4.55 & 2.78 & \underline{1.42} & \textbf{1.33} \\
    & 50m & - & - & - & - & - & 3.12 & \underline{1.57} & \textbf{1.51} \\
    & 100m & - & - & - & - & - & 3.31 & \underline{1.73} & \textbf{1.66} \\
    \bottomrule
    \end{tabular}
  }
\end{table}

\begin{figure}
  \centering
  \begin{subfigure}{0.31\textwidth}
    \centering
    \includegraphics[width=\textwidth]{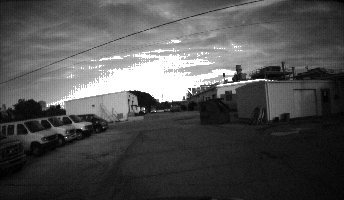}
    \caption{Reference image}
  \end{subfigure}
  \hfill
  \begin{subfigure}{0.31\textwidth}
    \centering
    \includegraphics[width=\textwidth]{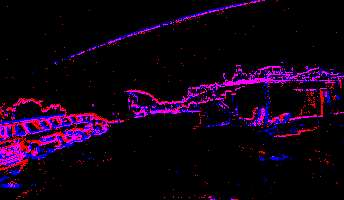}
    \caption{Events input}
  \end{subfigure}
  \hfill
  \begin{subfigure}{0.31\textwidth}
    \centering
    \includegraphics[width=\textwidth]{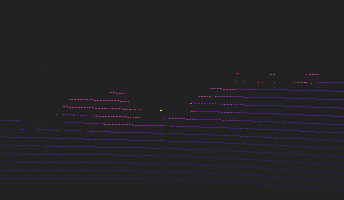}
    \caption{LiDAR input}
  \end{subfigure}
  \begin{subfigure}{0.31\textwidth}
    \centering
    \includegraphics[width=\textwidth]{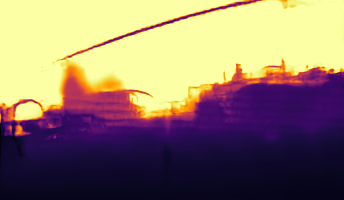}
    \caption{ALED\textsubscript{S}}
  \end{subfigure}
  \hfill
  \begin{subfigure}{0.31\textwidth}
    \centering
    \includegraphics[width=\textwidth]{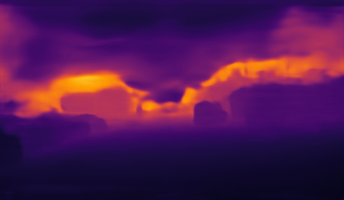}
    \caption{ALED\textsubscript{R}}\label{subfig:mvsec_cmp_r}
  \end{subfigure}
  \hfill
  \begin{subfigure}{0.31\textwidth}
    \centering
    \includegraphics[width=\textwidth]{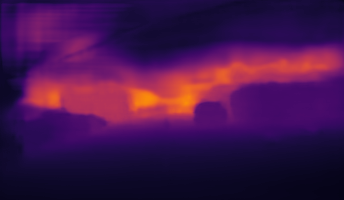}
    \caption{ALED\textsubscript{S\(\rightarrow\)R}}\label{subfig:mvsec_cmp_sr}
  \end{subfigure}
  \begin{subfigure}{0.31\textwidth}
    \centering
    \includegraphics[width=\textwidth]{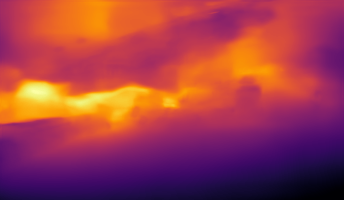}
    \caption{RAMNet~\cite{Gehrig2021CombiningEA}}
  \end{subfigure}
  \hfill
  \begin{subfigure}{0.31\textwidth}
    \centering
    \includegraphics[width=\textwidth]{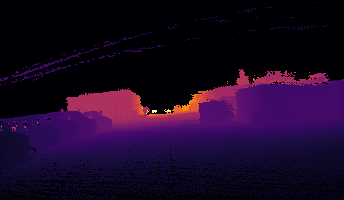}
    \caption{Ground truth}
  \end{subfigure}
  \hfill
  \begin{subfigure}{0.31\textwidth}
    \centering
    \vspace{10mm}
    \includegraphics[width=0.8\textwidth]{figures/mvsec_cmp/scale.png}
    \begin{tikzpicture}
      \node[] (black) {0m};
      \node[right=0.6cm of black] (pink) {50m};
      \node[right=0.3cm of pink] (yellow) {100m};
    \end{tikzpicture}
    \vspace{5mm}
    \caption{Scale}
  \end{subfigure}
  \caption{Qualitative results on ``Outdoor day 1'' from the MVSEC dataset with real data.}\label{fig:mvsec_cmp}
\end{figure}

Proposed ALED\textsubscript{S\(\rightarrow\)R} network greatly outperforms all the other approaches of the state of the art. Most impressive results are obtained with distant cutoff depths, where fewer LiDAR points are available: our network is still able to infer accurate depths, while reference methods show large errors. Compared to the frames+events EvT\textsuperscript{+} method of Sabater et al.~\cite{Sabater2022EventTA}, we improve the error by \(\{-2.7\%, 13.4\%, 31.3\%\}\) at minimum and by \(\{60.6\%, 58.8\%, 57.0\%\}\) at maximum for each of the \(\{10\text{m}, 20\text{m}, 30\text{m}\}\) cutoff distances respectively. Compared to the LiDAR+events method of Cui et al.~\cite{Cui2022DenseDE}, this improvement is of \(\{32.7\%, 17.4\%, 56.7\%\}\) at minimum and of \(\{59.7\%, 39.9\%, 79.1\%\}\) at maximum.

Qualitative results are presented in Fig.~\ref{fig:mvsec_cmp}. All three ALED variants produce results which are visually close to the ground truth for ground objects. Since the MVSEC dataset lacks ground truth depth for the sky, and since the rare elements to have a ground truth for this part of the image are close buildings, close trees, or power lines, the network cannot learn to derive correct depth estimations for the corresponding pixels, leading to the purple blobs in the upper parts of Fig.~\ref{subfig:mvsec_cmp_r}~and~\ref{subfig:mvsec_cmp_sr}. Only the \(S\) variant is able to predict accurate values for sky areas (as our SLED dataset contains valid ground truth depths for all pixels), but has more difficulties for ground objects due to the lack of fine-tuning. Between the \(R\) and \(S\rightarrow R\) variants, improvement can still be seen, for instance for the edges of the objects of Fig.~\ref{subfig:mvsec_cmp_r}~and~\ref{subfig:mvsec_cmp_sr}, which are less uneven in the \(S\rightarrow R\) variant. Finally, when comparing our results to RAMNet, we can clearly observe that our method provides in all cases more accurate depth maps, where object boundaries are more prominent, and where estimated depths are closer to the ground truth. This observation further demonstrates that the use of a LiDAR input --- even if very sparse --- is of great help for obtaining accurate dense depth maps.

\subsection{Associating a Depth to Each Event}
As stated in Sections~\ref{sec:intro}~and~\ref{sec:two_depths_per_event}, our goal is not only to estimate dense depth maps, but also to associate two depths to each event, allowing for their 3D reprojection and depth difference analysis. Evaluation of the depth association to each event on proposed SLED dataset is given in the ``Sparse depths errors'' column of Table~\ref{tab:results_sled}.

The sparse event-LiDAR fusion literature is limited to the method of Li et al.~\cite{Li2021Enhancing3L}. However, their approach only considers one depth per event, and is intended for a Road Side Unit (RSU) application (i.e., their event camera is fixed and evaluation is conducted on a specific dataset). Therefore, we decided to compare ourselves to a more naive (and faster) baseline: the Nearest Neighbor (NN) approach, where each event is given the depth of its closest LiDAR point. As the Nearest Neighbor approach can not infer correct depths for events which are too far from a LiDAR scan, and so as to provide a fair comparison, we only consider the events between the bottom and top LiDAR scans.

As displayed in Table~\ref{tab:results_sled}, in the ``before'' \(D_\text{bf}\) case, depending on the map and the cutoff distance, best results are shared between the NN approach and our ALED\textsubscript{S} network. These results can be explained by the fact that our network is more likely to commit large errors for events at the boundary of close objects, as it might estimate that they should be given the depth of the more distant background. On the contrary, the NN approach will always attribute the depth of the closest LiDAR point, and will therefore commit more frequent but smaller errors. In the ``after'' \(D_\text{af}\) case, despite this potential source of error, our network ALED\textsubscript{S} nearly always obtains the best results, as it has correctly learned the temporal propagation of the depths, a task which cannot be completed natively with the NN approach. We also remind here that these results are given for the parts of the image where LiDAR data is available: the NN method would not be able to derive correct estimations for the other parts of the image. Numerical results on each sequence of the dataset are given in Supplementary Material.

\subsection{Depth Difference}
We finally estimate the quality of our ``two depths per event'' approach through the depth change map, as presented in Section~\ref{sec:two_depths_per_event}. We perform two evaluations on our SLED dataset: (1)~the average error on the depth change map \(D_\text{af}-D_\text{bf}\) compared to the true depth changes, and (2)~the percentage of correctly classified events when using a difference threshold of 1m on the depth change map. Results of these evaluations are given in the ``Depth change map errors'' column of Table~\ref{tab:results_sled}. Numerical results on each sequence of the dataset, as well as some visual results, are given in Supplementary Material.

We can observe here that, despite a significant absolute error on the depth change maps, events can still be classified correctly, with a rate of success over 75\% on \textit{Town01}, and over 77\% on \textit{Town03}. We remind here that the ALED network is not trained on these depth change map and classification tasks. As such, we believe that, while even more accurate individual depth maps could improve both the depth change map and classification errors, further improvements could be brought by designing a network specifically dedicated to these tasks.

\section{Conclusion}

In this article, a novel learning-based approach for estimating dense depth maps from asynchronous LiDAR and event-based data has been proposed. A novel ``two depths per event'' notion has also been proposed, to solve the issue of events representing a change of depth. A synthetic multimodal dataset has also been recorded, to train and evaluate our method. Multiple evaluations on our synthetic and a real driving datasets have been performed to show the relevance of our contributions. In particular, on the MVSEC dataset, an improvement of up to 79.1\% compared to the current LiDAR-and-events state of the art has been achieved, on complex daytime and nighttime recordings.

In hindsight, further improvements could be brought to the method. Attention-based networks provide state-of-the-art results in numerous vision-based applications, and could improve the fusion of the event and LiDAR modalities. Making the network predict directly sparse depths for each event could also potentially provide better results for the depth to event inference. This could be achieved by using sparse convolutional networks~\cite{Messikommer2020EventbasedAS} for instance, and could be subject to future work. The use of the Event Volume with a fixed time window as the input representation for the events in our network could also be revised, as it can become ill-suited under large motions. A solution could be to use an alternative representation, such as TORE Volumes~\cite{Baldwin2021TimeOrderedRE}, or to use adaptive accumulation times using methods such as the one proposed by Liu and Delbrück~\cite{Liu2018AdaptiveTB}. Finally, the recording of a real dataset with a high resolution event camera could also be considered, to complete the possibilities offered by the low-resolution MVSEC dataset.

\bibliographystyle{splncs04}
\bibliography{bibliography}

\end{document}